\documentclass[11pt]{article}
\usepackage[margin=1in]{geometry}
\usepackage{iftex}
\usepackage[T1]{fontenc}
\usepackage{lmodern}
\ifPDFTeX
  \usepackage[utf8]{inputenc}
  \DeclareUnicodeCharacter{2212}{\ensuremath{-}}
\fi
\usepackage{amsmath,amssymb,graphicx,xcolor,fvextra}
\usepackage{longtable,booktabs,array,calc}
\usepackage{etoolbox,caption,float,placeins,xurl,needspace}
\usepackage{hyperref}
\hypersetup{colorlinks=true,urlcolor=blue,linkcolor=blue,hypertexnames=false,breaklinks=true}
\AtBeginEnvironment{longtable}{\small}
\providecommand{\tightlist}{\setlength{\itemsep}{0pt}\setlength{\parskip}{0pt}}

\makeatletter
\def\maxwidth{\ifdim\Gin@nat@width>\linewidth\linewidth\else\Gin@nat@width\fi}
\def\maxheight{\ifdim\Gin@nat@height>0.68\textheight0.68\textheight\else\Gin@nat@height\fi}
\makeatother
\setkeys{Gin}{width=\maxwidth,height=\maxheight,keepaspectratio}
\providecommand{\pandocbounded}[1]{#1}
\title{Agent Calibration under Deployment Change: Acceptance Contracts, Trajectory Repair, and Adaptation Costs}
\author{
\begin{tabular}{c}
Yaxiao Liu\textsuperscript{1}\quad Pengbo Liu\textsuperscript{1}\quad Yiwen Liu\textsuperscript{1}\\[0.3em]
Yihua Guan\textsuperscript{1}\quad Xingjian Li\textsuperscript{1}\quad Jiaxing Song\textsuperscript{2}\\[0.3em]
\small \textsuperscript{1}PwC China AI Center\quad \textsuperscript{2}Tsinghua University\\[0.3em]
\footnotesize \href{mailto:yaxiao.y.liu@cn.pwc.com}{\texttt{yaxiao.y.liu@cn.pwc.com}}\quad \href{mailto:rootliu@gmail.com}{\texttt{rootliu@gmail.com}}\\[0.3em]
\footnotesize \href{mailto:liupengbo@mails.neu.edu.cn}{\texttt{liupengbo@mails.neu.edu.cn}}\quad \href{mailto:202383049@uibe.edu.cn}{\texttt{202383049@uibe.edu.cn}}\\[0.3em]
\footnotesize \href{mailto:202311260036@mail.bnu.edu.cn}{\texttt{202311260036@mail.bnu.edu.cn}}\quad \href{mailto:23013142@mail.ecust.edu.cn}{\texttt{23013142@mail.ecust.edu.cn}}\\[0.3em]
\footnotesize \href{mailto:jxsong@tsinghua.edu.cn}{\texttt{jxsong@tsinghua.edu.cn}}
\end{tabular}
}
\date{November 25, 2026}
\begin{document}
\maketitle

\begin{abstract}

An agent can remain operational after a model, tool, or user-context
change while losing capabilities required at its destination. We define
agent calibration as qualification against a versioned deployment
contract. The contract specifies basic capabilities and end-to-end
reachability, technical and professional precision, and acceptance by
intended users. For each requirement, a source-to-destination map
records what must be preserved, transformed, added, or retired. Experts
supply benchmarks, positive and negative examples, annotated
trajectories, and cost limits. These instruments guide changes to
prompts and context, tool calls and other harness settings, or an
external compact policy while the foundation model remains frozen.
Candidate revisions are evaluated against fixed acceptance criteria
using independent evidence. We compare the costs of maintaining a
harness with and without an external policy, including annotation,
training, serving, review, and rework. Three diagnostic audits
illustrate the limits of aggregate scores. On a 100-item intent task,
each model solved cases the other missed. In a 60-record synthetic
financial reporting fixture, five records passed a numeric proxy but
failed the reporting contract. In a target-only WebMall snapshot, quota
interruptions affected 132 of 171 attempts. These observations motivate
paired reruns, checkpoint replay, and separate measurement of repaired
and newly failed cases. We specify controlled experiments to test
whether calibration increases accepted deliveries at a given cost.

\end{abstract}


\FloatBarrier
\section{1. Introduction: why deployment change requires
calibration}\label{introduction-why-deployment-change-requires-calibration}

Enterprise agents combine model behavior with organizational knowledge,
tools, execution policies, and recipient requirements. Replacing a
foundation model, adopting a post-trained version, moving a service, or
expanding into another business unit changes this combination. Even
versions from one model family need not expose a nested set of useful
behaviors. A source configuration can therefore lose previously
successful cases and gain others. For deployment, the relevant question
is whether the resulting agent meets the destination's requirements, not
whether one model is universally stronger.

A second change concerns professional meaning. An output may preserve a
correct number while assigning it to an incorrect period, field, unit,
or recipient template. Cross-section dependencies can make a local
formatting or semantic error consequential for an entire report. A third
concerns users: roles, permitted information flows, workflows,
preferences, and available resources can change independently of the
model. A correct answer that violates an authorization boundary or
cannot be consumed by the next system is not an acceptable delivery.

We call the specification, diagnosis, adaptation, and requalification of
an agent under these changes \emph{agent calibration}. This is
deployment qualification rather than probability calibration. Migration
describes a change; calibration determines what must remain invariant,
what must be transformed, and what evidence qualifies the new
deployment. A source agent is useful comparative evidence, but it is
neither ground truth nor a ceiling. Initial deployment can use
independent references without any source agent. Verification without
modification is a valid calibration outcome when the target already
qualifies.

Our argument has four parts. First, deployment changes create observable
requirements, including both invariants and authorized changes. Second,
these requirements become a versioned contract with basic, technical,
and user-context standards. Third, a bounded procedure combines
prompt/context adaptation, harness configuration, and optional external
policy training to repair gaps. These loci can interact, particularly
when preferences depend on the user's situation. Fourth, independent
evaluation distinguishes replacement differences, successful
qualification, incremental adaptation benefit, and lifecycle value.
These are distinct claims; none follows automatically from another.

We propose a source-to-destination requirement map organized by three
standard families and four supervision instruments; requirement-linked
trajectory repair for professional transformations; and a lifecycle
comparison of two adaptation routes. Frozen-backbone control, prompt
optimization, process rewards, runtime constraints, personalized
alignment, and closed-loop evaluation already have substantial
precedents {[}40{]}. Our testable claim is narrower: explicit
requirement disposition and dependency requalification may improve
accepted delivery under deployment change, beyond strong evaluation and
adaptation baselines, at a justifiable total cost. Existing results
diagnose gaps and evaluation failures; they do not establish that
incremental benefit.

Behavioral differences alone do not identify changed intrinsic knowledge
as their cause. Prompt interpretation, retrieval, sampling, tool
behavior, routing, and infrastructure also matter. We therefore require
recorded deployment versions and controlled interventions before
attributing changes to a model, harness, or learned policy.

\FloatBarrier
\section{2. Acceptance objectives: what must be
calibrated}\label{acceptance-objectives-what-must-be-calibrated}

\subsection{2.1. A versioned destination
contract}\label{a-versioned-destination-contract}

Let an agent consist of a foundation executor \(M_\phi\), configuration
\(\theta=(p,h)\), and optional external controller parameters \(\psi\).
Here \(p\) specifies prompt/context configuration, while \(h\) specifies
harness capabilities and resource settings. These components can be
adapted together. Its deployment environment supplies tools, data,
infrastructure, and user context. An execution produces an observable
trace \(\xi\), intermediate artifacts, a final artifact, and resource
consumption. We write the destination contract as

\[
C=(B,G,H,\boldsymbol{\tau},U,\mathcal{L}),
\]

where \(B\) is a mandatory test suite; \(G\) specifies real end-to-end
paths and postconditions; \(H\) contains hard predicates;
\(\boldsymbol{\tau}\) specifies minimum quality levels; \(U\) identifies
authorized user contexts; and \(\mathcal{L}\) gives resource limits.
Acceptance conditions are fixed before optimizing candidates. A
legitimate change of requirements creates a new contract version and is
reported separately from improvement under the old one.

Requirements are explicitly mapped across source and destination as
\textbf{Preserve, Transform, Add, or Retire}. Preserve protects
still-applicable knowledge and behaviors. Transform requires an approved
conversion with retained meaning. Add captures new obligations. Retire
removes an obsolete obligation only with authorization. This mapping
avoids penalizing a correct target template for differing from the
source, or silently treating a lost obligation as an upgrade.

A completed run must satisfy applicable hard predicates and
postconditions, meet its required quality thresholds, and remain within
its resource limits. PASS, FAIL, UNKNOWN, and NOT RUN are distinct
states. Missing evidence, mocks, and unexecuted paths cannot count as
passes. A permitted abstention can be a safe handling outcome while
remaining an incomplete delivery. Qualification of an agent is a
separate finite-suite judgment: all mandatory tests must pass, and
predeclared distribution-level criteria must hold on the evaluation
sample. Neither judgment is a universal guarantee beyond its scope.

The scope fingerprint records the model and provider identifiers,
decoding and routing configuration, harness and tool versions, data
snapshots, references, contract, checker, and resource policy. An
unverified provider identity is recorded as unverified. Contract
changes, checker corrections, and actual agent improvements have
separate histories.

Each requirement also carries a traceability record: its identifier and
authority, source-to-destination disposition, effective date, affected
artifact and dependent checkpoints, independent reference, acceptance
predicate, eligible supervision, and release consequence. This connects
a business change to an executable check without treating an expert
preference as ground truth. Conflicting or unsupported requirements
remain unresolved rather than being converted into confident reward
labels.

\subsection{2.2. Three standard families and four supervision
instruments}\label{three-standard-families-and-four-supervision-instruments}

\textbf{Basic calibration} covers essential competence and operational
reachability. It includes benchmark items, authentication, network and
storage access, tool invocation, and delivery to the intended endpoint.
A mock tool response does not qualify a production path. Infrastructure
censoring must remain visible in denominators.

\textbf{Technical calibration} covers professional precision and
artifact usability: terminology, formulas, units, temporal granularity,
code semantics, field identities, report components, and dependencies
between components. A report's structure can be semantically binding
even when all visible scalar values look plausible. Requirements come
from the applicable task and recipient; there is no universal national
report template implied by our examples.

\textbf{User-context calibration} covers the intended user's role,
background, use case, upstream and downstream information, permissions,
authorized preferences and values, and resource constraints. Values do
not authorize factual falsification or bypassing hard permissions. User
acceptance must be measured with appropriate recipients or explicit,
validated acceptance rules, rather than assumed from a generic judge
score.

User acceptance is conditional, not a permanent property of a person.
The same user may prefer different detail, tone, evidence, and
turnaround under different roles, audiences, tasks, environments, and
budgets. Subjective judgments may also be uncertain, inconsistent, or
shaped by the alternatives displayed. Record these conditions and
distinguish a preference change from an execution failure or an
authorized change of requirements. Different recipients' preferences
need not admit one universal ordering.

Hard versus soft is a separate distinction from the three families. A
professional formula, a privacy restriction, and a budget ceiling can
all be hard constraints. Optional presentation preferences can admit
trade-offs. Table 1 separates the standards from the optimization
instruments that help meet them.

\begin{longtable}[]{@{}
  >{\raggedright\arraybackslash}p{(\linewidth - 6\tabcolsep) * \real{0.1253}}
  >{\raggedright\arraybackslash}p{(\linewidth - 6\tabcolsep) * \real{0.2632}}
  >{\raggedright\arraybackslash}p{(\linewidth - 6\tabcolsep) * \real{0.3410}}
  >{\raggedright\arraybackslash}p{(\linewidth - 6\tabcolsep) * \real{0.2705}}@{}}
\caption{Three standard families, with many-to-many links to the four
instruments: Benchmark, Shots, Trajectory, and Cost. Hard constraints
may occur in any row.}\tabularnewline
\toprule\noalign{}
\begin{minipage}[b]{\linewidth}\raggedright
Standard family
\end{minipage} & \begin{minipage}[b]{\linewidth}\raggedright
Acceptance examples
\end{minipage} & \begin{minipage}[b]{\linewidth}\raggedright
Supervision and adaptation controls
\end{minipage} & \begin{minipage}[b]{\linewidth}\raggedright
Independent evidence
\end{minipage} \\
\midrule\noalign{}
\endfirsthead
\toprule\noalign{}
\begin{minipage}[b]{\linewidth}\raggedright
Standard family
\end{minipage} & \begin{minipage}[b]{\linewidth}\raggedright
Acceptance examples
\end{minipage} & \begin{minipage}[b]{\linewidth}\raggedright
Supervision and adaptation controls
\end{minipage} & \begin{minipage}[b]{\linewidth}\raggedright
Independent evidence
\end{minipage} \\
\midrule\noalign{}
\endhead
\bottomrule\noalign{}
\endlastfoot
Basic & Essential tasks; reachable delivery path & Benchmark cases;
tool-failure trajectories; retry budgets & Test results and observed
endpoint state \\
Technical & Units, periods, identities, report dependencies &
Positive/negative shots; checkpoint trajectories; benchmark predicates &
Reference-linked semantic and interface checks \\
User context & Role, use case, permission, recipient utility &
Role-conditioned shots, trajectories, and cost/utility priorities &
Authorized recipient assessment and access checks \\
\end{longtable}

\textbf{Benchmark} supplies mandatory cases and measurable objectives.
\textbf{Shots} supply positive and negative examples of acceptable and
unacceptable behavior. \textbf{Trajectory} supplies method structure,
checkpoints, branches, bounded loops, and observed success or failure.
\textbf{Cost} supplies resource ceilings, utility priorities, and
accounting rules. These are structured supervision assets, not four
interchangeable scalar hyperparameters. Their weights, sampling
frequencies, search budgets, and eligible action sets can be configured
during optimization. Mandatory gates cannot be relaxed by changing
reward weights. Skills are executable capabilities selected or
constructed by a harness, rather than a fifth standard or a replacement
for trajectory supervision.

\begin{figure}[!htbp]
\centering
\pandocbounded{\includegraphics[keepaspectratio,alt={Proposed reference architecture. Three standard families define acceptance; four supervision instruments guide three composable adaptation loci around a frozen foundation executor. Context-conditioned feedback can jointly change prompt/context, harness settings, and an external policy. Independent qualification remains outside the candidate's editable components. The architecture is a design proposal, not a measured system benefit.}]{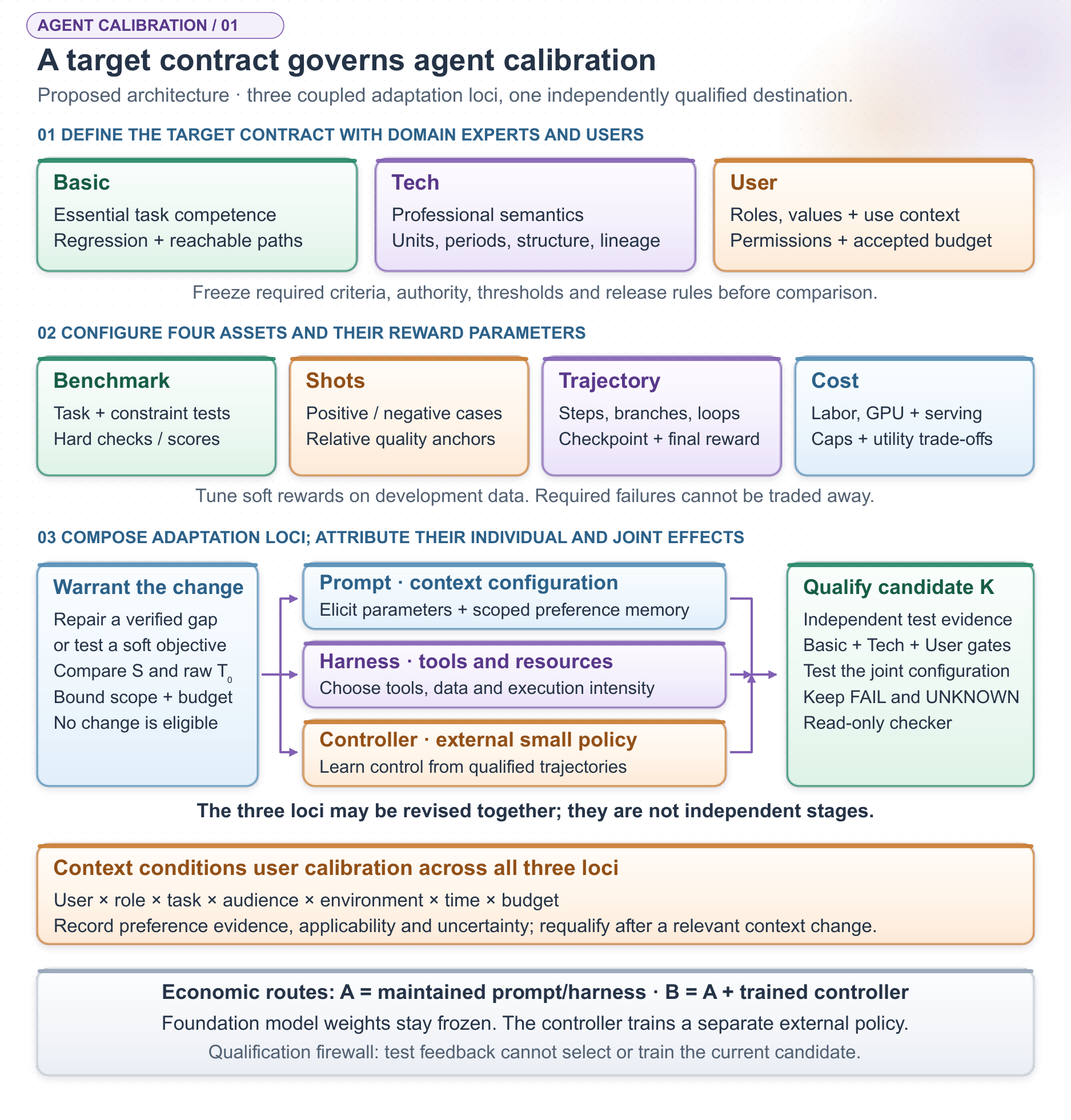}}
\caption{Proposed reference architecture. Three standard families define
acceptance; four supervision instruments guide three composable
adaptation loci around a frozen foundation executor. Context-conditioned
feedback can jointly change prompt/context, harness settings, and an
external policy. Independent qualification remains outside the
candidate's editable components. The architecture is a design proposal,
not a measured system benefit.}
\end{figure}

\subsection{2.3. Where the objectives differ across
applications}\label{where-the-objectives-differ-across-applications}

QA, Task, and Job are nonexclusive delivery profiles. QA emphasizes
supported claims, relations, and uncertainty. Task emphasizes an
artifact or action with observable postconditions. Job produces a
reusable capability, such as a compact policy exposed through a skill
API. These profiles allocate evaluation effort; they do not exempt any
agent from basic, technical, or user-context requirements. Appendix A
gives the profile-specific obligations.

For QA, a reference inventory of claims and relations is more auditable
than an unspecified assertion that information entropy is preserved.
Coverage, contradiction, unsupported additions, and missing relations
should be scored separately. Stronger target outputs can be authorized
improvements when independently supported. For Task, correct output is
insufficient without the required interface and postcondition. For Job,
both the supervision package and the resulting capability require
qualification; training rewards cannot replace an independent test
suite.

\FloatBarrier
\section{3. Related work and the remaining research
question}\label{related-work-and-the-remaining-research-question}

\subsection{3.1. Model-sensitive prompts and
harnesses}\label{model-sensitive-prompts-and-harnesses}

PromptBridge directly studies cross-model prompt transfer {[}1{]}.
HarnessDev and HarnessOpt-Bench make model-sensitive harness
construction and bounded optimization explicit {[}2, 3{]}. Beyond
Prompts evaluates tool-agent harness optimization {[}4{]}, while GEPA
uses reflective prompt evolution {[}5{]}. These precedents rule out a
novelty claim based merely on prompt or harness adaptation after a model
change. A credible comparison must include strong target-native
optimization with the same references, editable components, and
resources.

Workflow-Localized Mechanism Learning connects attribution, local
repair, and knowledge reuse in structured skills {[}6{]}. AEVAL provides
a precedent for deterministic workflow evaluation {[}7{]}. Consequently,
dependency-aware repair is a candidate mechanism to test, not an
established unique contribution of this paper. We ask whether an
explicit connection to versioned deployment requirements makes that
mechanism more effective or auditable across professional and user
changes.

GRACE {[}32{]} is a close structural precedent: it evolves persistent
instructions through a typed graph and local verification while keeping
the model, tools, and harness fixed. Our proposed editable scope is
broader, but that scope is not itself evidence of an advantage. A
matched comparison must distinguish instruction maintenance from changes
to tools, contracts, and external policies, and charge validation and
optimization compute consistently.

Recursive harness editing offers a concrete comparator for migration.
Harness\textsuperscript{2} {[}69{]} uses contrasting executions to propose and probe
edits. In Appendix E.7 of {[}69{]}, Figure 16 shows how a spreadsheet
aggregated at the wrong grain improved from 23.6 to 100.0 after the
harness specified grouping, derived fields, and input handling. In
Figure 18 of the same study, preserving visible edits through text
conversion improved the authors' score from 44.4 to 94.5. These are
same-task repairs under the authors' evaluator. A separate experiment is
needed to establish when further tuning helps and whether a repair
transfers. A behavioral difference can guide the next edit even when
neither execution establishes which artifact meets the destination
contract. Liu and Han {[}70{]} compare seven models across three coding
harnesses. On their hard45 tasks, the median task pass/fail flip rate
was 13\% both for unchanged reruns and for harness swaps across the
available run pairs. This makes an unchanged rerun an empirical control
for interpreting a paired repair. These studies address complementary
questions: whether an edit changes behavior, and whether the change is
distinguishable from sampling variability. Qualification adds a third
question: which destination obligation did the edit restore, and at what
delivery cost?

GUARD {[}71{]} directly tests the risk of editing a harness around a
frozen model. Its gate weighs gains against newly incurred losses on
visible tasks. Across five domains (236 test tasks), the authors report
macro accuracy of 65.81\% for the original harness, 66.46\% for GUARD,
and 56.11\% when the group winner is promoted unconditionally. The large
contrast is with unconditional promotion. A recipient may also require a
particular obligation to hold regardless of gains elsewhere. This
motivates comparing the two admission rules on the same proposed edits.

\subsection{3.2. Professional artifacts, context, and runtime
constraints}\label{professional-artifacts-context-and-runtime-constraints}

FinReportBench evaluates institution-grade financial reports, and Finch
covers spreadsheet-centered finance and accounting workflows {[}8, 9{]}.
CI-Work addresses contextual integrity in enterprise agents {[}10{]}.
Learning Personalized Agents from Human Feedback and TriAlign address
personalization and its relationship to truth consistency {[}11, 12{]}.
These works already emphasize concerns beyond generic academic task
accuracy. Our distinction is a common qualification procedure for
changed deployments, not the assertion that industrial or user-context
evaluation was previously absent.

AgentSpec specifies customizable runtime constraints {[}13{]}. Runtime
enforcement is complementary to adaptation: it can prevent or detect
forbidden behavior, whereas an adaptation policy may learn how to
complete a task within those constraints. A policy does not earn
permission to violate a rule merely because it predicts high utility.

APeB tests agent personalization using underspecified requests and noisy
behavioral histories {[}46{]}. This supports evaluating context
selection rather than assuming that more history improves service. Its
consumer setting does not validate enterprise role alignment, and its
data release remains subject to privacy review in the reviewed version.
In our protocol, inferred preference is defeasible evidence; explicit
role permissions remain authoritative.

\begin{longtable}[]{@{}
  >{\raggedright\arraybackslash}p{(\linewidth - 4\tabcolsep) * \real{0.2576}}
  >{\raggedright\arraybackslash}p{(\linewidth - 4\tabcolsep) * \real{0.3712}}
  >{\raggedright\arraybackslash}p{(\linewidth - 4\tabcolsep) * \real{0.3712}}@{}}
\caption{Closest precedents and testable distinctions. The right column
contains open hypotheses, not demonstrated advantages or priority
claims.}\tabularnewline
\toprule\noalign{}
\begin{minipage}[b]{\linewidth}\raggedright
Closest research line
\end{minipage} & \begin{minipage}[b]{\linewidth}\raggedright
Existing contribution relevant here
\end{minipage} & \begin{minipage}[b]{\linewidth}\raggedright
Question left for this proposal
\end{minipage} \\
\midrule\noalign{}
\endfirsthead
\toprule\noalign{}
\begin{minipage}[b]{\linewidth}\raggedright
Closest research line
\end{minipage} & \begin{minipage}[b]{\linewidth}\raggedright
Existing contribution relevant here
\end{minipage} & \begin{minipage}[b]{\linewidth}\raggedright
Question left for this proposal
\end{minipage} \\
\midrule\noalign{}
\endhead
\bottomrule\noalign{}
\endlastfoot
Prompt and harness optimization {[}1, 2, 3, 4, 5{]} & Adapt model-facing
configuration & Does contract-linked supervision improve joint
destination acceptance under matched access and cost? \\
Local workflow repair {[}6, 7{]} & Attribution, reusable repair,
workflow checks & When does dependency revalidation improve reliability
or reduce cost relative to full revalidation? \\
Scoped instruction evolution {[}32{]} & Typed graphs and local
verification with fixed execution infrastructure & Does
requirement-linked repair help when dependencies or deployment
components change? \\
Repair validity {[}33{]} & Oracle-checkable fabricated failures in a
deterministic micro-lab & Can a repair warrant reject unsupported edits
while retaining useful repairs? \\
Finance and contextual evaluation {[}8, 10{]} & Professional output and
information-flow criteria & Can one versioned procedure handle
interacting technical and user changes? \\
Frozen-executor adaptation {[}14, 41{]} & Train external control
policies or executable-harness editors & Which learned intervention
transfers and reduces lifecycle burden beyond maintained harnesses? \\
Rubric feedback and verifier reliability {[}22, 25, 37, 39{]} & Explicit
criteria guide learning; reward quality and checker reliability can
diverge & Does evidence-linked supervision improve independent
acceptance without increasing false acceptance or unnecessary
abstention? \\
On-policy correction under an adapted harness {[}36{]} & Round-scoped
expert correction recovers weak-model performance that full-trajectory
imitation degrades & Does repair scoping linked to a failed checkpoint
reproduce this asymmetry under professional deployment changes? \\
Closed-loop evaluation {[}40{]} & Atomic evidence, versioned
aggregation, evaluator audits, hard invariants, repair and regression &
What does explicit requirement mapping and dependency requalification
add under deployment change? \\
Adaptive process rubrics {[}42, 47, 48{]} & Task- or step-conditioned
feedback and process credit & Can feedback adapt without drifting from
destination obligations or hiding its cost? \\
Learned and revised rubrics {[}65, 66, 67, 68{]} & Learned weights,
personalized criteria, anchored revision and held-out certification &
Does profile-to-configuration adaptation improve actual deployment
acceptance beyond evaluator fit? \\
\end{longtable}

\subsection{3.3. Learning around a frozen foundation
model}\label{learning-around-a-frozen-foundation-model}

RLPrompt learns discrete prompting policies for a frozen downstream
model {[}15{]}. Learning to Control LLM Agent Harnesses with Offline
Reinforcement Learning is a particularly close precedent for a trainable
harness controller {[}14{]}. Agent Lightning separates execution from
agent training, while agent distillation transfers behavior to smaller
models {[}16, 17{]}. These methods support the feasibility of peripheral
learning but do not establish its deployment benefit in our setting.

Harness-R1 trains a separate engineer with SFT and GRPO to edit
executable harnesses, using fresh reruns of a frozen target agent for
rewards {[}41{]}. This directly supports learning around fixed
foundation weights, but its adaptation-time editor is distinct from a
controller invoked during each task. RRSI regularizes iterative harness
editing with edit budgets, leakage checks, and cost/complexity pruning
{[}43{]}. Both are relevant controls; neither establishes that a runtime
controller reduces our full labor and serving costs.

Offline advantage-weighted regression provides one training route
{[}18{]}. DPO optimizes preferences and GRPO uses relative group rewards
{[}19, 20{]}; neither generates an independent factual reference.
Process supervision and rubric rewards offer richer feedback than a
final binary outcome {[}21, 22, 23{]}. Reward hacking and impossible
rubric research demonstrate why optimizing such feedback still requires
checker qualification {[}24, 25{]}. Structural trajectory selection
further motivates inspecting the quality of the path as well as its
outcome {[}26{]}.

AdaPRA adapts process rubrics while inspecting workspace evidence
{[}42{]}; ARCO co-evolves a policy and rubric model {[}47{]}. ORSD
selects outcome-aligned rubric bundles and uses privileged training
context to shape token-level GRPO credit {[}48{]}. The latter two update
the acting policy, so they are not direct demonstrations of
frozen-executor calibration. Adapting their credit mechanisms to an
external controller is a proposed experiment, and outcome correlation
alone does not establish correct process attribution.

Imitation itself can be the failure mode. After a harness has been
evolved for a weaker model, fine-tuning that model on complete expert
trajectories degraded all seven enterprise tasks tested, by 4 to 30
points, reproduced across two model families, while on-policy correction
that rewrote only the failed rounds preserved the model's adapted
planning {[}36{]}. The same imitation was beneficial on an unevolved
harness (average test success 29.2\% to 35.5\% in their Qwen runs), so
the result bounds when trajectory supervision helps rather than ruling
it out.

CASE {[}35{]} is an adjacent example of learning evidence-seeking
workflows through trajectory supervision followed by agentic
post-training of a compact policy. It motivates a trainable peripheral
capability, but its medical setting and trained policy do not establish
frozen-executor calibration, enterprise transfer, or our proposed
acceptance rate.

\subsection{3.4. Limits of automated evaluation and
transfer}\label{limits-of-automated-evaluation-and-transfer}

The rubric-guided RL survey {[}39{]} organizes explicit criteria across
reward construction, process supervision, and policy updates, while
identifying semantic drift and linguistic reward hacking. It is a
synthesis, not new evidence that a deployment-calibration architecture
works. We use its separation of criteria, evaluation, and aggregation to
specify the interfaces in Section 4.1. The contribution sought here is
the connection between changed deployment obligations, inspectable
repair, independent qualification, and cost; neither structured rewards
nor expert criteria are new by themselves.

Closed-Loop Agent Evaluation is a closer integration precedent {[}40{]}:
it combines atomic evidence, deterministic versioned aggregation,
evaluator auditing, hard invariants, and repair with component
regression. Its business-analysis study supports evaluating the
measurement pipeline, but does not validate production releases of
proposed fixes. We therefore do not claim the evaluation loop as novel.
The incremental question is whether explicit requirement mapping and
transformation dependencies improve adaptation under changed
deployments, beyond that loop.

Criteria-exposure experiments show why a stronger verifier score need
not mean better task completion: access to misaligned success criteria
can increase measured success while reducing instruction-faithful
success {[}45{]}. Enterprise requirements should remain visible to
agents; qualification instances, reference answers, and exploitable
checker implementation details need exposure controls. Concealing
legitimate obligations would test a different problem.

Reference-free judges can be too generous, and user-simulated agent
assessment can misrepresent quality {[}27, 28{]}. Agreement between a
candidate checker and a stronger reference has limited diagnostic power
of its own: across medical, legal, and finance benchmarks judged by
frontier-LLM references, higher verifier agreement did not consistently
identify the training verifier that produced the best post-trained
model, beyond screening out very weak verifiers {[}37{]}. That study
concerns robustness to reward noise during optimization; it did not test
reward hacking or exploitable bias, and its reference judges were
themselves LLMs. It lowers the cost of starting optimization with an
imperfect supervision signal; it does not replace the independent
holdout qualification that Sections 4 and 7 require. A comparison
between two wrong answers remains a comparison between wrong answers.
Deterministic checks, independent sources, blinded expert adjudication,
and explicit unknown states are therefore necessary wherever the
application permits them. Continual optimizer evaluation and
cross-benchmark generalization studies also motivate testing adaptation
across changes rather than assuming that accumulated edits transfer
{[}29, 30{]}. The literature supports individual components and exposes
their failure modes; it does not establish the efficacy of their
integration proposed here.

Phantom Guardrails {[}33{]} shows in a deterministic micro-lab that
harness optimizers can invent failures and add unnecessary fixes. We
therefore require a claimed defect repair to identify an applicable
failure and its evidence. A separate soft-quality improvement or
preference probe needs a declared objective, comparator, and budget,
rather than an invented failure; no change remains a legitimate outcome.
This result motivates a control; it does not estimate how frequently
fabrication occurs in enterprise deployments. Evidence-bound
gateway-path provenance {[}34{]} separately motivates verifiable routing
and fallback records. Such records establish execution provenance within
their stated trust assumptions, not semantic correctness of an answer.

The remaining question is incremental value beyond existing integration:
whether requirement disposition, professional dependency repair, and
context authority improve accepted delivery under matched access, cost,
and checking. Section 7 separates Experiment A (destination acceptance
against strong target-native and closed-loop baselines), Experiment B
(trajectory and dependency mechanisms), and Experiments C and D
(transfer and lifecycle value). The proposed organization is not
evidence of superiority; a simpler established method may suffice.

\subsection{3.5. Interactive preference elicitation and candidate
search}\label{interactive-preference-elicitation-and-candidate-search}

Candidate selection is established prior art, not by itself a new
calibration algorithm. CUPID {[}50{]} uses preference comparisons for
online model selection; LISTEN {[}51{]} elicits multi-objective
preferences through language. TextGrad {[}52{]} and DSPy {[}53{]}
provide mechanisms for improving compound systems and prompts. In
creative generation, PromptCharm {[}54{]} supports mixed-initiative
refinement, while APPO {[}55{]} directly optimizes prompts through
repeated user choices among generated images. These precedents motivate
a low-burden interface for users who can recognize suitable results
without specifying professional generation parameters.

The research question here is narrower: can preference-guided search
over prompts, permitted evidence, plans, and harness configurations
produce reusable trajectories that remain qualified after deployment
change? Neither a preferred image nor a preferred enterprise answer
certifies all intermediate steps. Our proposed extension combines
observable choice histories with versioned constraints, execution
provenance, dependency checks, and subsequent transfer evaluation. It
does not claim that stochastic generation intrinsically prevents
overfitting or that image-generation findings establish enterprise-agent
reliability.

Recent work narrows the remaining gap. Online LLM Personalization
{[}56{]} studies pairwise selection from fixed persona prompts;
Personalize at Test Time {[}57{]} learns user-specific image
preferences. These support preference estimation, but do not establish
transfer of enterprise execution methods. AdaptArena {[}58{]} separates
inferred preferences from successful web execution. IMPLIED {[}59{]}
shows why fixed implications from feedback to unchosen actions can be
misleading. SRD {[}60{]} converts retrospective experience into
prospective supervision, while HarnessBandit {[}61{]} schedules
multi-harness policy training. The latter two update model parameters;
neither directly validates our proposed frozen-foundation external
controller. Our contribution is therefore the deployment qualification
protocol connecting these mechanisms, rather than invention of
preference learning or trajectory training.

Rubric learning is also direct prior art. Support Vector Rubrics (SVR)
learns global criterion weights with prompt-conditioned selection from
preference data {[}65{]}; it does not infer a changing per-user weight
profile. PARL generates personalized rubrics from user history and
trains its rubric generator with GRPO {[}66{]}. Its static preference
setting and cold-start limitations leave temporal change open, although
it already evaluates cross-user and cross-task generalization. Thus
neither learning rubric weights nor personalized rubric generation is
our novelty claim. RubricOpt further revises semantic rubrics under
evidence checks and fixed anchors for DPO {[}67{]}; RubricWorld uses
before/after grading transitions, fresh held-out certification, cost
accounting, and rejected-update fallback {[}68{]}. These precedents also
rule out claiming a verification-and-rollback loop itself as new. Our
prospective comparison concerns their extension to actual FM/harness
changes, execution receipts, destination contracts, and changing user
contexts; evaluation-rubric improvement alone does not establish
delivery improvement.

Multi-objective and elicitation work clarifies implementation choices.
LambdaMix conditions jointly trained internal adapters on supplied
preference weights {[}62{]}, rather than estimating weights or training
an external controller. OGC-DPO selects response-pair/objective queries
for multi-objective alignment {[}63{]}; its gradient access and training
compute differ from a black-box frozen executor. ELICITBENCH ties
elicitation to decision value and interaction burden {[}64{]}, but uses
simulated transactional users. These results motivate separate tests of
preference inference, accepted delivery, and full cost, rather than
assuming that more personalized scores or more questions improve all
three.

These precedents motivate composable intervention loci rather than a
fixed sequence. Preference elicitation and prompt selection
{[}51,55,56{]} inform prompt-level configuration; DSPy's composed
pipelines {[}53{]} connect prompt and workflow optimization.
Harness-R1's adaptation-time editor {[}41{]} and offline harness RL's
runtime controller {[}14{]} occupy different parts of the harness and
controller loci. Agent distillation supports learning tool behavior
{[}17{]}, while model--harness incompatibility {[}36{]} cautions against
assuming additive gains. Our proposed connection is shared, qualified
evidence with context and configuration provenance, evaluated under
fixed acceptance constraints and lifecycle budgets. Its integrated
transfer benefit remains an empirical question.

\FloatBarrier
\section{4. Candidate calibration
procedure}\label{candidate-calibration-procedure}

\subsection{4.1. Compile expert methods into inspectable
supervision}\label{compile-expert-methods-into-inspectable-supervision}

A domain expert supplies a method contract: the task decomposition,
required evidence, valid alternative branches, bounded iterations,
checkpoint predicates, and report-component requirements. The observed
trace records what an agent actually did. They are not interchangeable.
Supervision consists of observable operations and artifacts; it does not
require access to a model's private reasoning chain.

A checkpoint annotation identifies its inputs and provenance, expected
output, applicable rule, success and failure evidence, severity, and
dependent artifacts. Absolute rewards need anchored scales; relative
preferences need comparable tasks, contexts, and reference versions. A
partial failure may block an entire report, invalidate one component, or
permit a preauthorized degraded delivery. That consequence is part of
the annotation. Final report refinement is linked to component
requirements and user examples, rather than treated as an unconstrained
stylistic score.

Expert methods can evolve over time. A registry should therefore
distinguish the evidence's effective date from the method's version
date. Comparing contemporaneous and later methods can generate useful
contrastive examples, but later information unavailable at the task date
must not become a hidden input or ground truth for an earlier decision.
Temporal holdouts test whether a learned policy handles genuinely new
cases and requirements.

The registry stores versioned benchmarks, references, examples, method
graphs, actual traces, annotations, and available skills. Selection
records applicability and exclusions. Mere volume does not establish
quality, representativeness, or coverage. Annotation disagreement and
uncertainty remain visible rather than being collapsed into a single
confident reward.

Four distinct objects make this compilation operational. The
\textbf{contract} states what the destination accepts; a \textbf{rubric}
instantiates those requirements for a task and its observable
checkpoints; a \textbf{verifier} assesses the evidence against each
criterion; and a \textbf{reward rule} converts assessments into
optimization feedback. Rubrics operationalize the three standard
families using the four supervision instruments; they are neither a
fourth standard nor a fifth instrument. Appendix D gives the minimum
interfaces. A generated rubric is a proposal requiring coverage and
applicability checks, not authority to rewrite the contract.

Adaptive feedback may refine checkpoint questions or evidence requests
while preserving requirement coverage. AdaPRA documents a persistent
criterion that conflicted with the task metric {[}42{]}, motivating
explicit review and retirement of invalid feedback. Locking acceptance
does not preserve a known-broken verifier: an authorized correction
creates a new checker version, reevaluates baseline and candidate
consistently, and separates measurement changes from agent improvements.
Candidate optimizers cannot authorize that correction themselves.

Each criterion carries a requirement ID, scope and effective version,
hard/soft status, an evidence locator, and an assessment rule. Record
PASS, FAIL, UNKNOWN (insufficient or conflicting evidence), and NOT-RUN
(verification not executed) separately. Applicability exclusions need
explicit authorization rather than a passing score. A mandatory UNKNOWN
or NOT-RUN prevents a complete qualification claim. For an approved
abstention, the scoped abstention requirement can pass while the
requested report remains undelivered; both outcomes must be recorded.
Otherwise, a policy could appear reliable merely by declining useful
work.

Reflection must reference an observable check, not merely assert that
checking occurred. Link the requirement and proposed action to its
runtime receipt, result artifact, comparator, and criterion verdict. For
a repair, retain the before/after artifacts and the affected checks.
Missing execution is NOT-RUN; an executed check with insufficient or
conflicting evidence is UNKNOWN. Neither a fluent rationale nor a
claimed improvement substitutes for the receipt. Generate the
human-readable audit summary from this same record, retaining raw
observations when reward weights change. Such traceability makes a claim
inspectable; it does not itself establish that the verifier is correct.

For a financial artifact, the presence of five headings can be checked
mechanically but cannot establish correct period allocation, coherent
cross-section totals, or authorized disclosure. Those obligations
require distinct evidence and checks. Rubric decomposition should
preserve these dependencies instead of rewarding surface completeness
repeatedly. An expert-approved example anchors interpretation; it does
not establish factual truth without supporting data.

\subsection{4.2. Diagnose a gap before selecting a
repair}\label{diagnose-a-gap-before-selecting-a-repair}

We first qualify the environment and evaluate a source agent \(S\), an
unchanged target configuration \(T_0\), and, later, a frozen adapted
candidate \(K\); each prospective arm in Section 7 is one such
candidate, and Figures 1 and 3 use this label. Initial deployment can
omit \(S\). At the item level, source-only successes identify losses,
target-only successes identify gains, and both-fail cases identify
unresolved needs. These transitions are defined for a named endpoint and
contract; execution success and strict contract success may produce
different tables.

A gap record links a failed requirement to evidence and candidate
causes: missing knowledge, inadequate evidence selection, an incorrect
transformation, a tool or interface fault, an unreachable path, or a
user-context mismatch. A local diagnosis is a hypothesis until a
controlled change tests it. For example, a quota interruption is an
operational failure rather than proof of weaker reasoning. A provider
label in a report is not sufficient evidence of the executor actually
used.

Candidate repairs can change instructions, examples, context
construction, retrieval, tool wrappers, executable transformations,
sandbox access, or an external controller. The chosen edit must fit the
permitted action space and resource ceiling. Equivalent semantic
checkpoints, rather than identical tool sequences, align source and
target workflows. Requiring the target to outperform the source at every
intermediate step would reject valid alternative paths and is not our
criterion. Whole-trajectory replacement has a documented failure mode
under a harness adapted to the current model {[}36{]}; scoping the edit
to the diagnosed failure is an empirical precaution, not a stylistic
preference.

For an observed gap, record the expected and observed states at the
first affected checkpoint. Use signed deviations when the reference
defines units and direction, missing or conflicting evidence when it
does not, and a categorical verdict for a broken path. A retrieval edit
might change a date filter, replace a source, or supply a
counterexample. Log the actual parameter change, the evidence it added
or removed, and the downstream predicates it could affect. This gives a
prompt or context edit a testable prediction without assigning an
arbitrary distance to its wording.

Distinguish the first observable violation from its suspected origin.
Wang et al.~{[}72{]} studied 95 failed trajectories. In 37 of 88
Who\&When traces the model-annotated candidate origin preceded the
human-labeled decisive step. A retrospective label suggests where to
intervene. For example, if a report assigns a transaction to the wrong
month, replay from before period assignment using the same input rows.
Compare a date-rule edit with an unchanged rerun under the same tool
snapshot. Record whether the first wrong subtotal and the final
reconciled total both recover. If the prefix cannot be reconstructed,
retain the attribution as unresolved.

\subsection{4.3. Repair dependencies, not only the final
presentation}\label{repair-dependencies-not-only-the-final-presentation}

Consider a report that consumes weekly totals but requires ten-day or
monthly periods. Weekly totals alone generally do not identify daily
values. If raw daily records exist, the correct path is to retrieve,
validate, and reaggregate them. Otherwise, an explicitly approved
estimation method may use relevant indicators and disclose uncertainty.
Uniform allocation is an assumption, not an exact recovery. Temporal
disaggregation and reconciliation have established statistical methods
{[}31{]}; an LLM should not invent missing precision.

A changed period transformation invalidates dependent aggregates,
ratios, tables, narrative conclusions, and final checks. The harness
must rerun or revalidate this affected closure. WorkRover provides an
adjacent repair-selection control: with repair proposals fixed,
compatible edit selection reduced cross-task regression from 7.0\% to
0.7\% relative to applying all edits, at higher model-token cost
(normalized 1.00 versus 0.59) {[}38{]}. This concerns environment
repair, not its separate verifier-to-turn credit mechanism or evidence
of our deployment efficacy. An unaffected independent branch can
continue where the method contract permits. A mandatory failed or
unknown checkpoint blocks dependent release; optional degradation
requires prior authorization. If the dependency graph is missing or
unreliable, the fallback is conservative full revalidation or
escalation, not silent reuse of stale artifacts.

\begin{figure}[!htbp]
\centering
\pandocbounded{\includegraphics[keepaspectratio,alt={Proposed dependency-aware period repair. Weekly totals do not determine daily observations. Raw records or an approved estimation route must establish the intermediate representation before aggregation into the recipient's periods. Changes invalidate dependent tables and conclusions; unknown mandatory evidence blocks release. This is an illustrative method graph, not an empirical result.}]{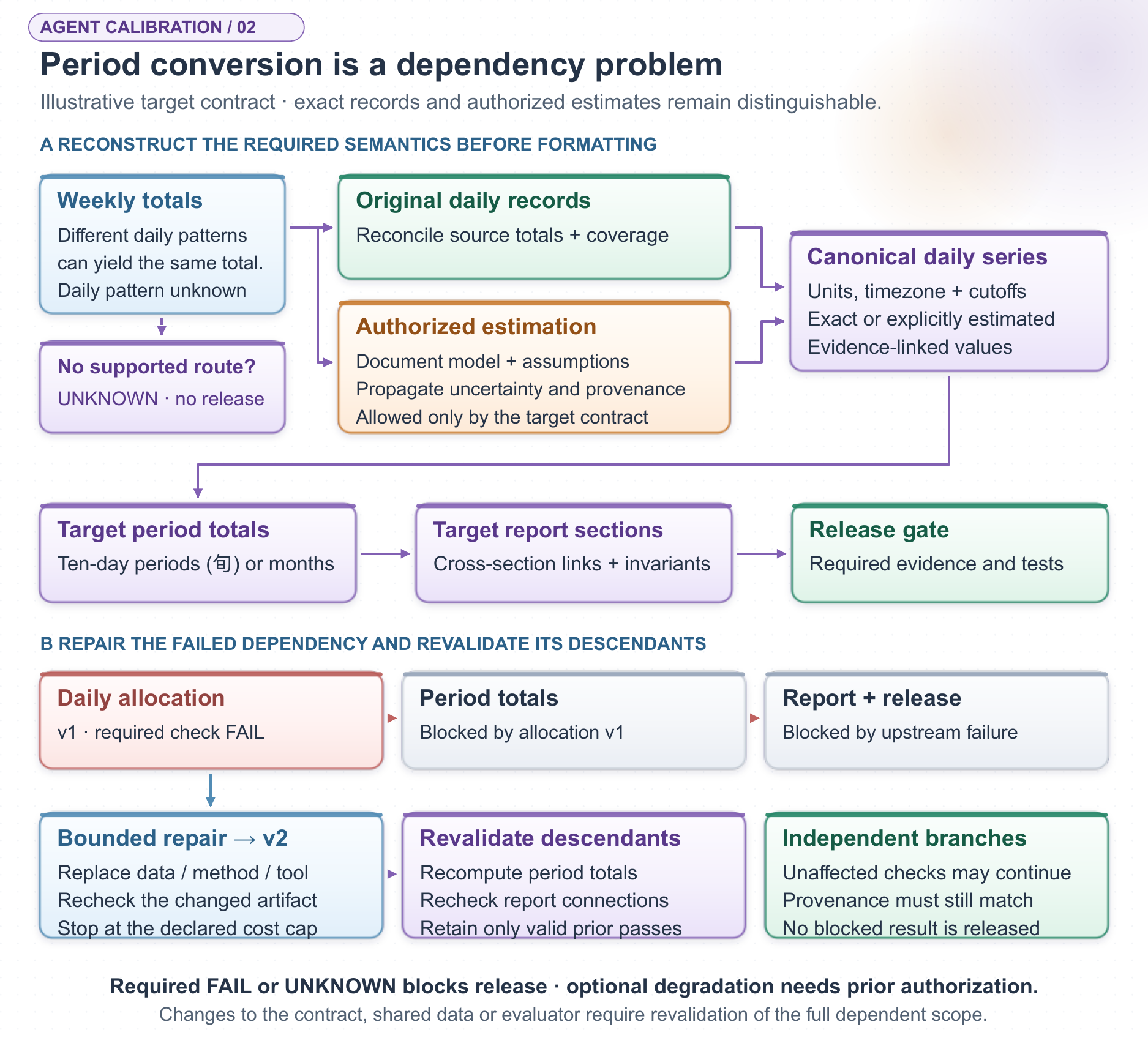}}
\caption{Proposed dependency-aware period repair. Weekly totals do not
determine daily observations. Raw records or an approved estimation
route must establish the intermediate representation before aggregation
into the recipient's periods. Changes invalidate dependent tables and
conclusions; unknown mandatory evidence blocks release. This is an
illustrative method graph, not an empirical result.}
\end{figure}

\subsection{4.4. Optimize within a fixed contract, then
requalify}\label{optimize-within-a-fixed-contract-then-requalify}

A candidate objective can combine benchmark performance, example
preferences, checkpoint and outcome rewards, and resource penalties. Let
\(z\) denote the recorded user, role, task, audience, environment, and
resource context. Schematically,

\[
\max_{p,h,\psi}\;\mathbb{E}_{(x,z)\sim D_{\rm cal}}[R_{\lambda(z)}(\xi;C,z)]
\quad\text{subject to fixed admissibility rules and }\mathcal{L}.
\]

The soft weights \(\lambda(z)\) and the search or training budget are
configurable, context-conditioned controls. They do not redefine \(C\).
Preference weights can be elicited or estimated with uncertainty; a
single choice does not identify them uniquely. Freeze the conditioning
rule and update policy before evaluation, even when an authorized
runtime interaction remains available. Optimization feasibility or a
high reward is not qualification; independent tests still determine
release. Selection among feasible candidates should expose trade-offs
rather than bury mandatory failures in an average score.

One candidate-generation strategy separates anchored soft criteria from
current contextual preferences. For an eligible candidate, let \(r_a\)
score the former and \(r_u\) the latter under recorded, comparable
scales. Explore \(R_\alpha=\alpha r_a+(1-\alpha)r_u\), with
\(\alpha\in[0,1]\), while enforcing the same mandatory gates for every
setting. Endpoints and intermediate weights generate contrasting search
objectives; mixing scores does not guarantee interpolated answers or a
diverse candidate set. An apparently context-only objective still obeys
factual, professional, permission, and resource obligations. Long-term
memory is not automatically authoritative, and immediate feedback is not
automatically reliable: both require scope and provenance. Freeze scale
anchors, prevent duplicate criterion credit, and retain the separate
scores and actual configurations. Internal adapter mixing {[}62{]} is a
distinct realization; these configuration-level experiments do not
assume equivalent policy behavior. Weighted candidates are a
configurable search mechanism, not an additional acceptance dimension or
supervision instrument.

\textbf{When to repair, tune, or stop.} A structural failure can make
the next score uninterpretable. If the required retrieval did not
execute, first obtain a tool receipt and verify that it returned the
intended records. If execution already works, measure the gap directly.
Stop editing when all required checks pass and the remaining objectives
are within their agreed tolerances. Run independent qualification on the
unchanged or repaired candidate in either case.

For a remaining gap, give each criterion an explicit role during search.
A \textbf{gate} is a contract obligation that every eligible candidate
must satisfy. A \textbf{maintenance floor} protects an already attained
level on a soft criterion. A \textbf{phase weight} directs attention to
a remaining objective among eligible candidates. These roles can apply
within any of the three standard families. For example, while tuning
explanation quality in a financial report, the accounting period must
stay correct, and previously attained evidence coverage must stay within
its declared tolerance. A weighted average alone could reward a clearer
explanation that drops supporting records.

Specify each floor's units, direction, reference configuration, and
maximum acceptable loss before comparing edits. For a sampled metric,
use a paired non-inferiority check against that fixed reference on
development cases. An inconclusive interval warrants more evidence
within budget or an unresolved verdict. Keeping the reference fixed
prevents successive small losses from accumulating. Optimize coupled
criteria together if one edit moves both. Record the retained
configuration and outstanding gap when no feasible edit fits the budget.

ProRubric makes criteria noncompensatory within each scoring dimension,
while still aggregating dimensions by a weighted sum {[}44{]}. GEAR
models prerequisite and activation relations in a rubric graph and
propagates probabilistic support before reward aggregation {[}49{]}.
This reward-dependency graph differs from the artifact dependencies in
Section 4.3, which determine what a deployment change invalidates and
must requalify. Soft dependency-aware rewards may retain useful learning
signals under uncertainty; mandatory release predicates remain gates
across dimensions. Stylistic quality cannot compensate for a forbidden
disclosure or invalid period conversion. Preserve criterion-level
verdicts and test aggregation independently of rubric wording.

GRPO and DPO are optimization mechanisms, not independent truth
estimators. Group-relative rewards can rank several invalid candidates,
and a preferred DPO response can still violate a mandatory requirement.
Retain absolute gate verdicts alongside relative preferences; compare
pairs under the same task, available evidence, role, and contract
version. Exclude unresolved pairs or route them for adjudication rather
than manufacture a preference. Failure examples remain useful
supervision when their violated requirements are explicit. Changing soft
weights requires development evaluation; changing an acceptance
requirement creates a new contract version and a new qualification
claim.

The bounded procedure is:

\begin{enumerate}
\def\labelenumi{\arabic{enumi}.}
\tightlist
\item
  Freeze the contract, applicability map, independent checker, resource
  policy, and calibration/development/test partition.
\item
  Verify real environment paths; diagnose \(T_0\), and \(S\) when a
  source agent exists, using named endpoints and references.
\item
  Record the reason for intervention. A defect repair links an
  applicable requirement and failed checkpoint to inspectable evidence.
  An improvement or elicitation trial instead names an adjustable soft
  objective or unresolved preference, comparator, and budget; it need
  not invent a failure. Retaining the configuration remains an eligible
  decision. Record the change set and cost.
\item
  Revalidate the affected dependency closure and mandatory gates;
  escalate when prerequisites are unknown or the budget is exhausted.
\item
  Select a candidate on development evidence. Freeze its policy,
  artifacts, and allowed runtime repair rules.
\item
  Evaluate it once under the declared held-out protocol and release only
  within the qualified scope. Report unresolved failures.
\end{enumerate}

Iteration is bounded; success is not guaranteed. A valid endpoint is an
explicit unmet requirement with an escalation record. The optimizer may
edit candidate validators for operational use, but the qualification
checker and its references remain read-only and independently versioned.

A held-out run can perform bounded runtime repair if that rule was
frozen beforehand. Its test feedback cannot update a reusable policy and
still count as untouched evaluation; doing so requires a new holdout.
Source traces may be used offline for diagnosis. A protocol that
consults the source online is a different deployment regime and must
charge both agents' resources.

\begin{figure}[!htbp]
\centering
\pandocbounded{\includegraphics[keepaspectratio,alt={Proposed qualification protocol. Source versus unchanged target diagnoses replacement differences; adapted versus unchanged target estimates incremental adaptation under a common contract. Qualification is an additional decision against fixed gates. Calibration and development can guide edits; held-out evaluation uses a frozen policy and cannot silently feed reusable training.}]{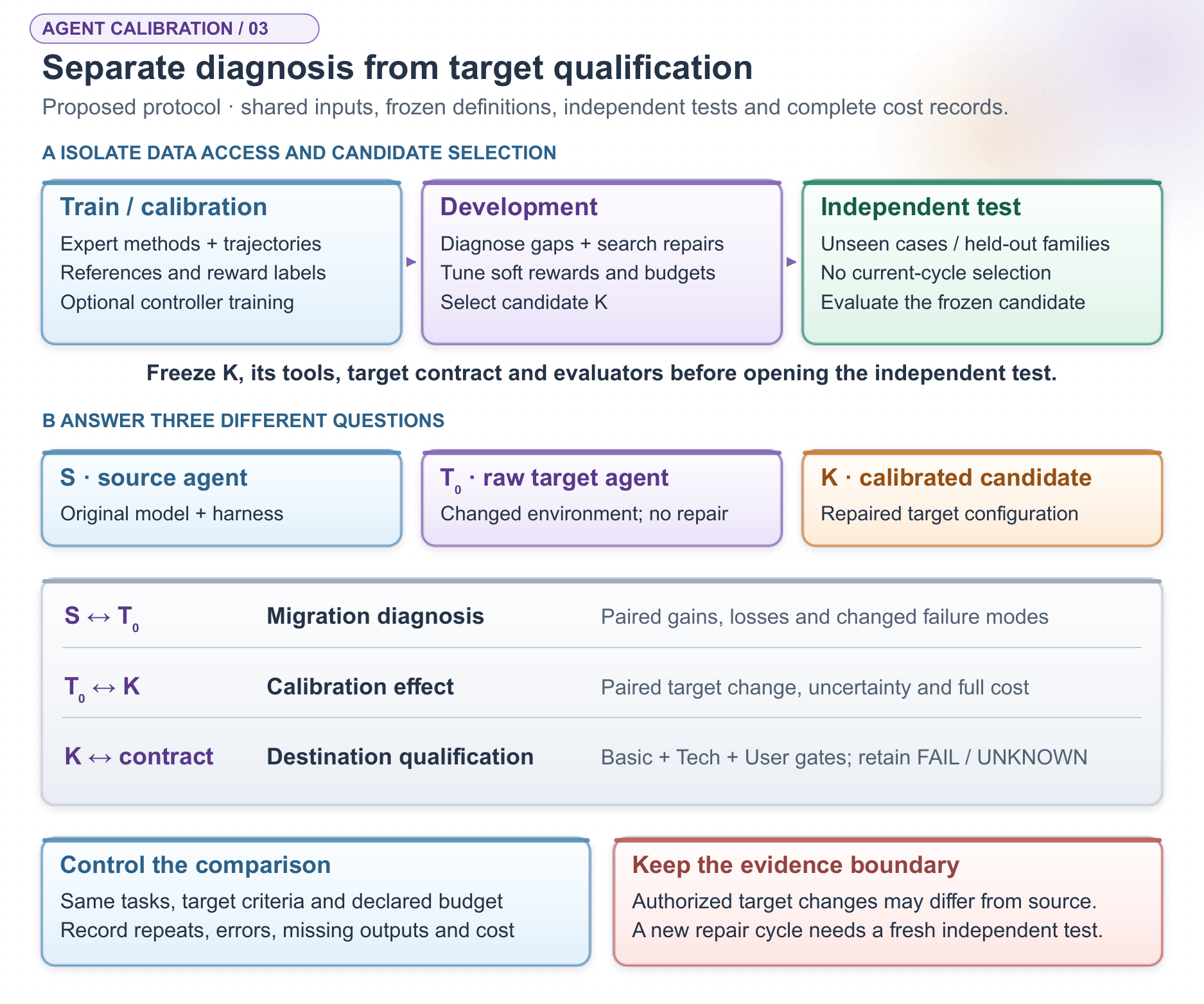}}
\caption{Proposed qualification protocol. Source versus unchanged target
diagnoses replacement differences; adapted versus unchanged target
estimates incremental adaptation under a common contract. Qualification
is an additional decision against fixed gates. Calibration and
development can guide edits; held-out evaluation uses a frozen policy
and cannot silently feed reusable training.}
\end{figure}

\subsection{4.5. Preference-guided candidate-and-trajectory
search}\label{preference-guided-candidate-and-trajectory-search}

\begin{figure}[!htbp]
\centering
\pandocbounded{\includegraphics[keepaspectratio,alt={Proposed preference-guided calibration architecture. (a) Natural-language intent is translated into candidate executable configurations and outputs; human selection and edits guide refinement, subject to basic, technical, and user acceptance gates. (b) Configuration and execution evidence is independently verified and curated into demonstrations, explicit preference pairs, or reward-bearing rollouts. (c) An optional learned controller configures a frozen foundation model and tools; held-out quality, transfer, and cost evaluation determines qualified reuse. Configuration parameters are distinct from learned weights, and an unselected alternative is not automatically a rejected example. The diagram summarizes a proposed mechanism, not a demonstrated training result.}]{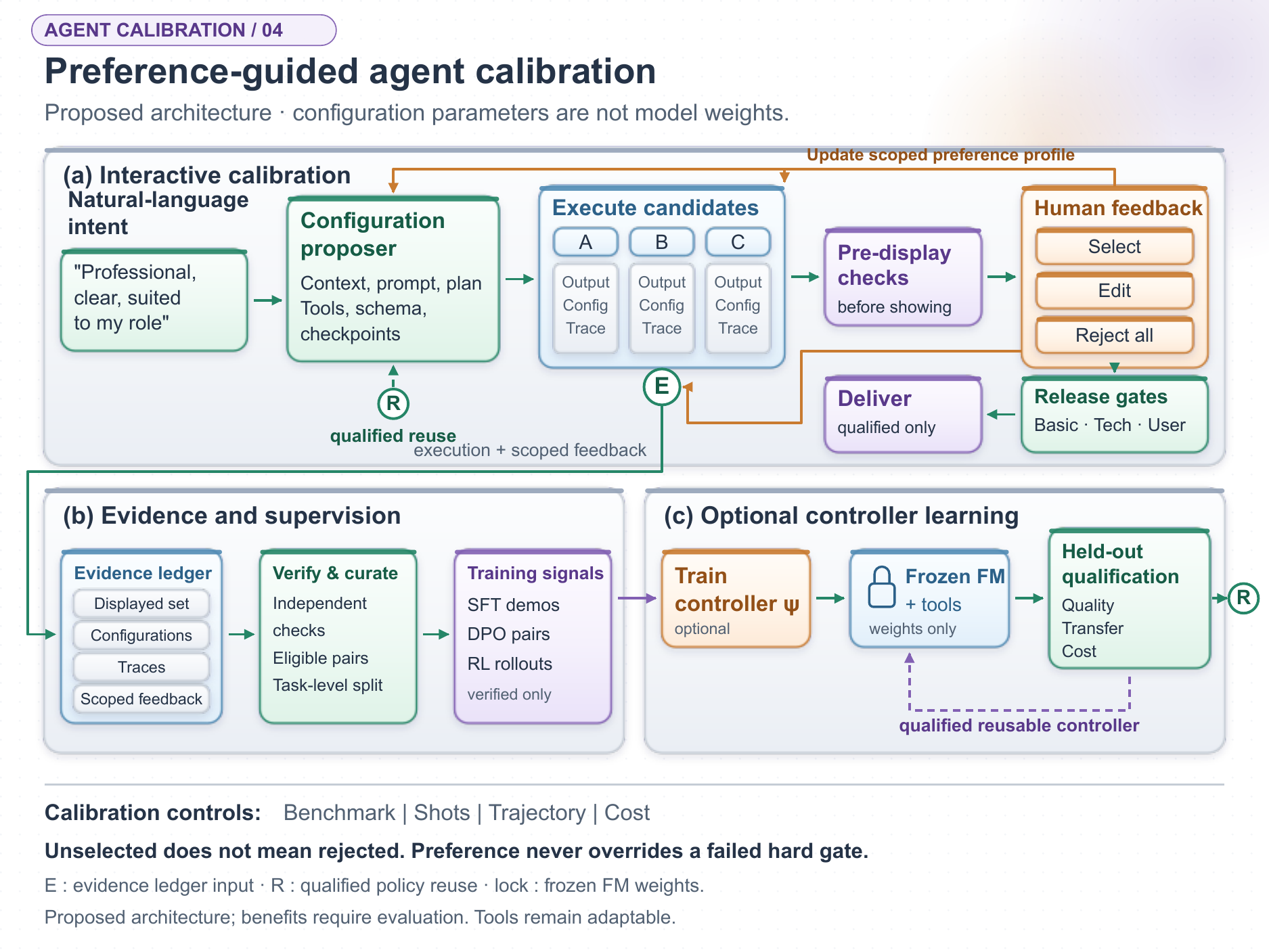}}
\caption{Proposed preference-guided calibration architecture. (a)
Natural-language intent is translated into candidate executable
configurations and outputs; human selection and edits guide refinement,
subject to basic, technical, and user acceptance gates. (b)
Configuration and execution evidence is independently verified and
curated into demonstrations, explicit preference pairs, or
reward-bearing rollouts. (c) An optional learned controller configures a
frozen foundation model and tools; held-out quality, transfer, and cost
evaluation determines qualified reuse. Configuration parameters are
distinct from learned weights, and an unselected alternative is not
automatically a rejected example. The diagram summarizes a proposed
mechanism, not a demonstrated training result.}
\end{figure}

In Figure 4, scoped feedback can revise a soft preference profile
(Section 4.6), which proposes changes to executable configurations
during interaction; neural weight training is an optional downstream
stage for qualified evidence. This evidence dependency does not impose
three independent adaptation phases: a previously qualified controller
can propose prompt/context and harness changes jointly. Pre-display
checks filter candidate presentation, while the final three-dimensional
gates determine delivery. Connector E routes execution evidence and
separately scoped feedback into the ledger. Both pass through
verification and curation before training. Paired R connectors allow
qualified policy reuse by the configuration proposer only after held-out
evaluation. The lock denotes frozen foundation-model weights, not
immutable tools.

Some requirements are easier to recognize than articulate. A user
creating a short video may choose a composition and request a change
without knowing camera movement or lighting terminology. Similarly, a
report recipient may recognize an appropriate explanation without
specifying the complete prompt, evidence join, and tool sequence. We
propose an optional interaction loop: generate alternatives, select or
edit, regenerate, and retain qualified reusable methods. Calibration can
remain unobtrusive in the interface while its constraints and provenance
remain explicit internally.

Separate two evolving objects. The fixed contract contains
non-negotiable factual, technical, permission, and resource
requirements. A provisional preference model summarizes
context-dependent choices within that feasible region. Preferences may
be discovered or constructed during interaction, rather than reveal a
pre-existing scalar utility. If a choice changes an actual acceptance
requirement, version the contract and re-evaluate affected candidates;
do not silently change the evaluation target.

For each round, propose configurations containing context, prompt,
authorized data sources, plan, tools, and checkpoints. Distinguish
multiple samples from one configuration from deliberate configuration
changes. Execute in the permitted sandbox and perform all applicable
pre-display checks. A mandatory user assessment that requires this
interaction remains pending or NOT-RUN; passing pre-display checks does
not yet qualify the delivery. Failed or unresolved candidates may be
shown as explicitly labeled diagnostics, never as accepted deliveries.
Present alternatives with randomized positions and comparable
formatting; allow ties, rejection of all candidates, localized edits,
and optional reasons. Model-generated alternatives can reduce the
expertise needed to specify a solution, but their search space still
reflects model and proposer biases.

Record the complete displayed set, exposure order, selected and rejected
candidates, localized feedback, user role, contract version, model and
source versions, configuration, actual tool trace, checkpoint outcomes,
and time and compute costs. An observable execution trace is not a
reconstruction of hidden model reasoning. A selection means conditional
preference among displayed alternatives; it does not prove correctness,
global optimality, or the quality of every step. Assign checkpoint
rewards only from scoped verification or explicit step-level feedback.
Controlled component changes are needed to attribute improvements to
prompts, tools, or plans.

Maintain separate records for observed feedback, inferred preferences,
and verified obligations. An unselected candidate is not necessarily
rejected: retain explicit pairwise choices separately from unseen,
skipped, tied, or unexamined alternatives. A local edit should update
its identified scope, not reward every preceding action. When several
configurations explain a choice, retain competing hypotheses and use
bounded contrasting candidates or clarification to distinguish them
{[}59{]}. Preference confidence cannot override a failed hard
obligation.

A counter-probe deliberately varies an unlocked soft axis against a
provisional inference: after a brighter image is selected, a dimmer
alternative can test whether brightness was preferred or merely
accompanied the least objectionable option. Hold other axes fixed where
feasible, randomize presentation, and label the interaction as a probe.
Never reverse a mandatory or explicitly locked requirement. Record the
proposed contrast, exposure, response, context, and human/computational
cost; silence is not confirmation. One contrast does not identify a
latent preference, especially when attributes interact or the user's
context changes. Store provisional, explicitly confirmed,
contract-locked, and expired states separately. Repeat a probe only when
its expected information value justifies attention and delay; faster
apparent convergence is not the sole endpoint. Objective-specific query
selection {[}63{]} and decision-sensitive elicitation {[}64{]} motivate
comparisons with random querying, direct clarification, and no probe,
with human and compute costs reported separately.

Promote experience through three separate gates: session acceptance,
independent execution qualification, and held-out reuse. Store the
applicability scope, evidence supporting or contradicting reuse,
expiration conditions, and rollback target. If retrospective critique is
distilled into a controller, the student input must contain only
information available before the decision; future outcomes may supervise
learning but must not leak into its input {[}60{]}. The proposed
external-controller implementation remains to be evaluated.

Initially retrieve and adapt qualified templates with frozen foundation
weights. Promote a template only with an applicability scope and
validation on separate tasks. With sufficient qualified data, train an
external proposal or control policy. DPO {[}19{]} requires comparable
preferred and rejected actions under the same conditioning; preference
for a final answer cannot automatically label an unrelated controller
action sequence. GRPO {[}20{]} likewise needs grounded group rewards and
must not manufacture a successful winner from an entirely invalid group.
Neither learning method supplies an independent truth reference.

This mechanism adds a preference-elicitation path to the four
instruments: benchmarks retain hard tests; selected and rejected
examples enrich shots; verified executions enrich trajectories; and cost
bounds include human attention, candidate generation, verification, and
later training. Stop on acceptance, budget exhaustion, or unresolved
constraints, reporting which occurred. Session improvement and
subsequent autonomous transfer are separate outcomes. Repeated user
assistance is not zero-shot performance.

\subsection{4.6. From choices to a versioned rubric-weight
profile}\label{from-choices-to-a-versioned-rubric-weight-profile}

Building on learned and personalized rubrics {[}65,66{]}, the
interaction can update a context-conditioned profile of soft rubric
weights, not only select the current answer. Keep the criterion
definitions and scale anchors fixed while estimating which trade-offs
explain comparable choices. A simple candidate estimator is a
regularized pairwise preference model over criterion-score differences,
with nonnegative normalized weights and an explicit tie or unresolved
state. This is a possible implementation, not an identified model of
human utility. Correlated criteria, narrow candidate coverage,
inconsistent judgments, and missing evidence can leave several weight
profiles equally plausible. Retain uncertainty or decline an update; do
not treat an unobserved contrast as a negative label.

The profile records user and context scope, rubric and scale versions,
eligible feedback IDs, estimated weights and uncertainty, applicable
generation controls, expiration conditions, and a rollback target. A
requirement's authority comes from the contract: even an
ordinary-language user instruction may establish a mandatory or locked
condition. Such a condition is not a weight to be reduced because a
candidate looks attractive. Conversely, an expert's stylistic suggestion
need not become a hard gate. Changes to criterion meanings,
applicability, or mandatory predicates require an authorized
contract/rubric revision and fresh comparison, separately from fitting
weights over unchanged criteria.

\begin{longtable}[]{@{}
  >{\raggedright\arraybackslash}p{(\linewidth - 4\tabcolsep) * \real{0.2230}}
  >{\raggedright\arraybackslash}p{(\linewidth - 4\tabcolsep) * \real{0.3869}}
  >{\raggedright\arraybackslash}p{(\linewidth - 4\tabcolsep) * \real{0.3901}}@{}}
\caption{Separate profile estimation, executable adaptation, policy
training, and requirement revision. An improved preference fit does not
certify a configuration or authorize a new acceptance
standard.}\tabularnewline
\toprule\noalign{}
\begin{minipage}[b]{\linewidth}\raggedright
Updated object
\end{minipage} & \begin{minipage}[b]{\linewidth}\raggedright
Admissible evidence and action
\end{minipage} & \begin{minipage}[b]{\linewidth}\raggedright
Required check before reuse
\end{minipage} \\
\midrule\noalign{}
\endfirsthead
\toprule\noalign{}
\begin{minipage}[b]{\linewidth}\raggedright
Updated object
\end{minipage} & \begin{minipage}[b]{\linewidth}\raggedright
Admissible evidence and action
\end{minipage} & \begin{minipage}[b]{\linewidth}\raggedright
Required check before reuse
\end{minipage} \\
\midrule\noalign{}
\endhead
\bottomrule\noalign{}
\endlastfoot
Soft weight profile & Comparable choices or scoped edits; estimate
trade-offs within fixed criteria & Held-out preference prediction,
uncertainty, context scope and scale stability \\
Prompt/context or harness mapping & Profile suggests candidate
instructions, evidence selection or tool settings & Actual execution
receipts and independent destination qualification \\
External controller parameters & Qualified demonstrations, action
preferences or grounded rollouts & Frozen-executor comparison on new
tasks, users and declared environments \\
Requirement or criterion meaning & Explicit authorized revision with
preserved history & Versioned coverage review, baseline rescore and
affected requalification \\
\end{longtable}

A learned weight is not an executable instruction. A proposer must map
the profile to candidate prompt/context and harness changes, whose
actual effects are checked again. For example, a stronger preference for
auditability may suggest source-linked tables, additional retrieval, or
a verification tool; none is guaranteed to improve auditability. Store
the mapping and its execution evidence, including rejected updates and
no-change decisions. Version the profile and freeze its update rule
before evaluation. Feedback collected during an authorized test-time
session can support session adaptation, but cannot also be recycled as
untouched evidence of subsequent autonomous transfer.

This governance loop can operate without neural post-training. Only
later, when independently qualified records justify it, may an external
policy learn the profile-to-configuration mapping or trajectory
decisions while the foundation executor remains frozen. A cohort or
industry profile is a possible scoped prior, not a substitute for the
current user's requirements. Learning such priors requires
permission-compatible data, isolation of individual histories, and a
separate unseen-user evaluation; sharing sensitive trajectories is not
implied. Testing whether this loop improves accepted delivery beyond
learned-rubric baselines, rather than merely fitting choices more
closely, is the prospective claim.

\FloatBarrier
\section{5. Three coupled adaptation loci and their economic
objective}\label{three-coupled-adaptation-loci-and-their-economic-objective}

\subsection{5.1. Prompt/context, harness, and external
policy}\label{promptcontext-harness-and-external-policy}

The three loci identify \emph{where} adaptation acts, whereas Basic,
Technical, and User identify \emph{what must be accepted}. Their
relationship is many-to-many. User calibration may jointly change
instructions, permitted evidence, tool intensity, and learned routing;
professional precision can require the same combination. Table 4
separates the editable objects and their supervision.

\begin{longtable}[]{@{}
  >{\raggedright\arraybackslash}p{(\linewidth - 4\tabcolsep) * \real{0.1370}}
  >{\raggedright\arraybackslash}p{(\linewidth - 4\tabcolsep) * \real{0.4120}}
  >{\raggedright\arraybackslash}p{(\linewidth - 4\tabcolsep) * \real{0.4510}}@{}}
\caption{Composable adaptation loci. The prompt and harness loci may use
search or a learned editor; only changes to an explicitly identified
policy constitute weight training. No locus uniquely corresponds to an
acceptance dimension.}\tabularnewline
\toprule\noalign{}
\begin{minipage}[b]{\linewidth}\raggedright
Adaptation locus
\end{minipage} & \begin{minipage}[b]{\linewidth}\raggedright
Editable object
\end{minipage} & \begin{minipage}[b]{\linewidth}\raggedright
Supervision and reusable artifact
\end{minipage} \\
\midrule\noalign{}
\endfirsthead
\toprule\noalign{}
\begin{minipage}[b]{\linewidth}\raggedright
Adaptation locus
\end{minipage} & \begin{minipage}[b]{\linewidth}\raggedright
Editable object
\end{minipage} & \begin{minipage}[b]{\linewidth}\raggedright
Supervision and reusable artifact
\end{minipage} \\
\midrule\noalign{}
\endhead
\bottomrule\noalign{}
\endlastfoot
Prompt locus & Structured instructions, context selection, examples,
plan proposals & Expert standards and explicit choices; scoped
configuration memory \\
Harness locus & Authorized tool and database selection, retrieval depth,
execution/check/retry policies & Executed trajectories, checkpoint
feedback, resource records; versioned harness \\
Controller locus & Weights of a proposal, routing, or specialized skill
model; FM weights stay fixed & Qualified shots and traces, justified
preferences and rewards; reloadable policy with interface contract \\
\end{longtable}

At the prompt locus, convert underspecified language into inspectable
candidate settings: terminology, audience, detail, evidence scope,
output structure, and permitted plan alternatives. Expert references and
candidate selection refine those settings. Discrete configuration
search, template retrieval, and memory updates are learning from
interaction in a broad sense, but must not be reported as neural
post-training. Memory stores the configuration, elicited context,
supporting and contradicting feedback, contract version, evidence
provenance, applicability, expiry, and rollback. A synthesized
trajectory is a proposed method until execution establishes its trace
and checkpoint evidence.

At the harness locus, the selected configuration is realized through
authorized tools and data. Search depth, evidence joining, sandboxed
operations, checks, and bounded repair may change together with the
prompt locus. A more detailed prompt cannot supply an inaccessible
database; a richer database can change the useful prompt and plan.
Evidence quantity is not automatically evidence quality. Tool
permissions and hard budget limits remain enforced independently of
preference scores.

At the controller locus, an external policy can predict prompt-level
settings, choose among harness-level actions, or implement a repeated
skill. Its state includes only information available at decision time,
with context and permission fields explicitly defined. Its outputs are
typed configurations or allowed actions, not assumed access to the
foundation model's internal reasoning. The phrase \emph{activation
layer} denotes this external interface, not modification of internal
activations. Generic instruction-following improvement, transfer, and
lower cost are outcomes to measure rather than consequences of using a
smaller model.

\subsection{5.2. Joint adaptation and qualified
training}\label{joint-adaptation-and-qualified-training}

Choose an admissible subset of \((p,h,\psi)\) for each warranted repair
or scoped improvement. Joint search, alternating updates, or a frozen
controller proposing configuration changes are all possible
implementations. None implies that the loci are statistically
independent or must be completed in order. A final-answer preference
supervises the displayed configuration as a whole; attributing credit to
one prompt, data source, or checkpoint requires additional evidence. Log
each change and revalidate its affected dependencies. Freeze the
resulting composite deployment for independent evaluation.

For user calibration, maintain context-conditioned preference hypotheses
rather than one immutable user score. A choice between a concise
executive report and a detailed analyst report can be explained by
audience, task, budget, or presentation bias. Compare candidates under
matched conditions, retain ties and rejection of all options, and seek
clarification or contrasting evidence when the hypotheses imply
different actions. A permission change requires authorized contract
revision; it cannot be inferred merely from liking an output. Test
same-user context shifts separately from transfer to new users.

A practical training sequence is qualified traces, behavior cloning or
supervised learning, and then an optional test of preference
optimization or reward-based updates. It is a data-readiness sequence,
not an obligatory order of adaptation across the three loci. Supply
nontrained baselines with the same retrievable shots and memory so that
extra information does not masquerade as a training gain. Offline
advantage-weighted training differs from online rollout-based RL. DPO
requires justified pairs under matched task, state, evidence, and
context; preferences for final artifacts cannot automatically label
distinct hidden controller actions. GRPO requires informative group
rewards; retain ties and all-invalid groups rather than manufacture a
valid winner. Reward and process models require independent validation.

The controller and harness must be qualified together: new tools alter
states and available actions, while prompt changes can alter the traces
on which the controller was trained. Whole-trajectory imitation may
disrupt an adapted workflow {[}36{]}. Compare localized corrections
against full-trace targets and check compatibility after a harness
change. This motivates a factorial experiment and incremental controller
comparisons, not an assumption of additive gains.

\subsection{5.3. Two maintained implementation
routes}\label{two-maintained-implementation-routes}

The first economic route maintains the prompt and harness loci without a
trained runtime controller. Deterministic converters, schema checks,
retrieval, prompt selection, tool adapters, sandbox policy, and bounded
reflection can address failures with explicit engineering. Such work is
often CPU-oriented but is not free: evaluation, search, retries, and
reflection may still invoke the frozen foundation model. An
adaptation-time learned editor also incurs construction costs even if it
is absent during serving. Human diagnosis, integration, and regression
review can dominate expenditure.

This route is attractive when rules are precise, changes are infrequent,
or failures have simple verifiable repairs. Its limits include growing
configuration interactions, repeated manual diagnosis, and brittle
case-specific patches. Those limits are hypotheses to measure in a
longitudinal study; they do not imply that repeated edits necessarily
justify training.

The second route adds a compact controller \(\pi_\psi\) at the third
locus, while retaining prompt and harness maintenance and keeping
\(M_\phi\) frozen. It can select evidence, compile context, choose a
tool, request a check, allocate a bounded repair step, or escalate. A
runtime controller, an adaptation-time editor, and a specialized skill
policy have different invocation patterns and must be metered
separately.

For repeated professional trajectories, a specialized policy
\(\pi_\chi\) can be distilled and exposed as a skill API. It needs
explicit input/output contracts, permitted data access, versioned
capabilities, and a fallback for unsupported cases. Evaluation must
include clean reload, independently checked postconditions, unseen task
families, and changed user contexts. A manufacturing planning skill, for
example, cannot inherit production qualification merely from fitting
expert demonstrations. Training the compact model is neither mandatory
for every calibration nor established by the offline prototypes reported
below.

\subsection{5.4. Compare lifecycle cost and accepted delivery
value}\label{compare-lifecycle-cost-and-accepted-delivery-value}

The comparison is \textbf{maintained harness} versus \textbf{maintained
harness plus trained policy}. Both continue to require maintenance,
monitoring, and requalification. Let \(j\) denote either route. Over
periods \(t=1,\ldots,T\), define

\[
TC_j(T)=F_j+\sum_{t=1}^{T}\left(M_{jt}+G_{jt}+O_{jt}+N_t c_{jt}+E_{jt}\right).
\]

Here \(F_j\) is initial setup; \(M_{jt}\) is maintenance, annotation,
and review labor; \(G_{jt}\) is training and search compute; \(O_{jt}\)
is fixed operation; \(N_t c_{jt}\) is serving cost for the declared
workload; and \(E_{jt}\) is otherwise uncounted remediation or failure
cost. Accounting categories must be exclusive. For example, an API
charge should not also be billed as an estimated underlying GPU rental,
and rework labor already in \(M\) should not be duplicated in \(E\).

GPU reporting should identify hardware, occupied hours, utilization or
allocation assumptions, and monetary conversion. Serving reports need
tokens, calls, tool charges, latency, and missing-meter coverage. Labor
reports need role, activity, measured time, and the valuation rule.
Faster delivery can release staff capacity without reducing payroll.
Report time saved, cash savings, throughput, and monetized benefits
separately.

For controller route \(C\) versus harness route \(H\), a discounted
value difference is

\[
\Delta V=-(F_C-F_H)+\sum_{t=1}^{T}
\frac{(C_{Ht}-C_{Ct})+(V_{Ct}-V_{Ht})}{(1+r)^t},
\]

where \(C_{jt}\) is the period cost in the preceding equation and
\(V_{jt}\) is independently valued accepted delivery. Compare routes
that satisfy the same contract before interpreting lower cost as an
improvement. Benefits should not be invented from reward increases.
Whether training pays off depends on recurring workload, reusable
failure patterns, annotation burden, maintenance saved, and new
model-specific regression risk. No current experiment estimates this
break-even point.

\FloatBarrier
\section{6. What the existing experiments
establish}\label{what-the-existing-experiments-establish}

\subsection{6.1. Evidence scope and
endpoints}\label{evidence-scope-and-endpoints}

The materials below are supplied experiment records and offline audits,
not a prospective controlled evaluation of the integrated method. In the
intent study, \(S\) denotes the source agent and \(T_0\) the unmodified
target. The experiment report describes T1 as a revised self-check
prompt, T2 as a retry after failure, and C-lite as keyword-based
post-processing of \(T_0\) outputs. Each study defines its own
interventions. None of these pilots implements the full procedure of
Sections 4--5. Table 5 separates the observations from the claims they
cannot support. Appendix B preserves additional denominators and
resource limitations. Overlapping files and rerendered summaries are not
additional experiments.

\begin{longtable}[]{@{}
  >{\raggedright\arraybackslash}p{(\linewidth - 6\tabcolsep) * \real{0.1868}}
  >{\raggedright\arraybackslash}p{(\linewidth - 6\tabcolsep) * \real{0.2811}}
  >{\raggedright\arraybackslash}p{(\linewidth - 6\tabcolsep) * \real{0.2836}}
  >{\raggedright\arraybackslash}p{(\linewidth - 6\tabcolsep) * \real{0.2484}}@{}}
\caption{Diagnostic evidence and claim boundaries. Endpoints and
independent units differ across rows and must not be
pooled.}\tabularnewline
\toprule\noalign{}
\begin{minipage}[b]{\linewidth}\raggedright
Material and unit
\end{minipage} & \begin{minipage}[b]{\linewidth}\raggedright
Audited or supplied observation
\end{minipage} & \begin{minipage}[b]{\linewidth}\raggedright
Supported interpretation
\end{minipage} & \begin{minipage}[b]{\linewidth}\raggedright
Not established
\end{minipage} \\
\midrule\noalign{}
\endfirsthead
\toprule\noalign{}
\begin{minipage}[b]{\linewidth}\raggedright
Material and unit
\end{minipage} & \begin{minipage}[b]{\linewidth}\raggedright
Audited or supplied observation
\end{minipage} & \begin{minipage}[b]{\linewidth}\raggedright
Supported interpretation
\end{minipage} & \begin{minipage}[b]{\linewidth}\raggedright
Not established
\end{minipage} \\
\midrule\noalign{}
\endhead
\bottomrule\noalign{}
\endlastfoot
Intent B-to-A, 100 labeled items & 72 both correct; 3 source-only; 7
target-only; 18 both wrong & Aggregate gain can coexist with lost
source-correct cases & Provider-specific causality; integrated
calibration efficacy \\
FinQA, repeated questions & R1 execution: 17 losses/12 gains; R2: 7
losses/3 gains & Endpoint-specific replacement differences; repair can
introduce failures & Independent cluster intervals; controlled
superiority \\
E3, 60 records from one synthetic family & 50 strict passes; 5
numeric-proxy-pass/contract-fail; 2 unparsed; 3 missing & Proxy and
fixture contract can disagree & Broad factual correctness; user
acceptance; calibration effect \\
WebMall Test, 57 tasks/171 attempts & 132 quota interruptions; 27 URL
passes; 25 URL-plus-JSON passes & Operational censoring and interface
qualification require separate reporting & Matched source-target
improvement; verified order completion \\
\end{longtable}

\subsection{6.2. Intent: improvement and regression can occur
together}\label{intent-improvement-and-regression-can-occur-together}

On the supplied B-to-A labels, \(S\) scores 75/100 and \(T_0\) scores
79/100. The paired table contains 72 joint successes, three source-only
successes, seven target-only successes, and 18 joint failures. Thus a
four-point aggregate increase coexists with three lost cases. The loss
rate is 3/100 over all items or 3/75 among source-correct items; both
denominators should be explicit. This is a supplied-label recount,
without independent verification of provider identity or gold validity.

T1 scores 80, T2 scores 78, and C-lite scores 79. Relative to \(T_0\),
T1 repairs two of 21 errors and introduces one error among 79 previously
correct cases. Its one-point net gain is descriptive, without a
justified significance or minimum-detectable-effect claim. C-lite has
identical predicted labels to \(T_0\) on all 100 items. This does not
imply identical full responses or costs. The separate A-to-B summary is
not a controlled reverse-direction pair. These observations support
reporting gains and regressions; they do not show that the proposed
calibration procedure works.

\subsection{6.3. FinQA: endpoint choice and repair accounting
matter}\label{finqa-endpoint-choice-and-repair-accounting-matter}

R1 contains 100 questions, three repeats, and four arms: 1,200 episodes.
R2 contains 100 questions, one repeat, and four arms: 400 episodes.
These are not 1,600 independent questions. On the execution endpoint,
replacement loses 17 and gains 12 cases in R1, and loses seven and gains
three in R2.

C-lite versus \(T_0\) gains eight and induces ten failures in R1, and
gains four and induces four in R2. Recovery specifically among original
source-to-target losses is 5/17 and 2/7. These restricted recoveries
cannot be subtracted from all induced failures to obtain an overall
effect. New failures among \(T_0\)-correct episodes are 10/190 and 4/60.
Strict success and execution success are different endpoints: the former
increases by one point in each supplied aggregate, while the latter
changes by approximately minus 0.67 points and zero.

The supplied R1 summary reports an execution difference interval for
C-lite minus \(T_0\) of approximately {[}-4.17, 2.31{]} percentage
points, and C-lite minus T2 of {[}0.35, 6.74{]}. The supplied report
describes 88 clusters; the mapping and reported intervals were not
independently reproduced. These intervals are retained as reported, not
recertified. Model metadata conflicts, ambiguous summary-file
provenance, incomplete manifests, uneven resource feedback, and
incomplete human-review records further limit attribution. R2 reports
376/400 internal-gate passes and 238/400 strict successes across all
four arms, including the source. If these totals refer to the same
records, at least 138 gate passes are not strict successes; the
row-level intersection was not verified. This aggregate discrepancy is
not a validated estimate of an independent checker's false-positive
rate.

\subsection{6.4. E3: a numeric proxy is not an artifact
contract}\label{e3-a-numeric-proxy-is-not-an-artifact-contract}

The E3 fixture comprises 60 records from one synthetic fact family,
spanning two profiles, two splits, five arms, and three repeats.
Recounting yields 50 strict passes, not the earlier summary's 52: 27/30
DEV and 23/30 Test. Five records pass the numeric proxy but fail the
fixture contract, two cannot be parsed, and three are missing. The five
disagreements involve two identity issues and three disputed date rules.
They are not evidence that five artifacts are comprehensively factually
correct but unusable. Conflicting date specifications also make checker
validity part of the diagnosis.

T0, C-lite, and T2 each achieve 4/6 Test passes; T1 achieves 5/6 and
\(S\) 6/6. Calibration specifications were empty or inactive in the
runner; feedback and budgets differed, and initial generations were not
shared. DEV and Test reuse one fact family. No controlled calibration
effect, independent generalization, user acceptance, or downstream
damage is established. The valid lesson is narrower: professional
artifact qualification needs explicit identities, periods, and interface
rules in addition to numeric proxies.

\subsection{6.5. WebMall: availability and consumable output are
separate}\label{webmall-availability-and-consumable-output-are-separate}

In the primary target-only Test snapshot, 57 tasks generate 171
attempts. There are 132 quota interruptions and 39 executed attempts.
URL scoring passes 27; strict JSON and URL scoring jointly pass 25.
Accordingly, URL success is 27/171 (15.79\%) over all attempts or 27/39
(69.23\%) among executed attempts. Joint success is 25/171 (14.62\%) or
25/39 (64.10\%). The two URL-only passes illustrate a measured interface
discrepancy; no observed downstream loss is available, and URL scoring
is not an independent cart or order postcondition.

Three of the 60 planned tasks are absent from this snapshot and are
disclosed separately from its 171-attempt denominator. Nine of the 13
executed tasks pass URL scoring on all three repeats, and four fail on
all three. Calibration, DEV, and Test phases use different tasks under
the same prompt; phase names do not create an intervention comparison.
Another snapshot overlaps 132 keys and cannot be added as independent
evidence. The average time over all attempts is also distorted by early
quota interruptions.

\begin{figure}[!htbp]
\centering
\pandocbounded{\includegraphics[keepaspectratio,alt={Diagnostic observations with separate denominators. Intent counts are supplied-label outcomes for 100 items; E3 counts are 60 records from one synthetic family; WebMall counts are 171 target-only attempts with 132 quota interruptions. Panels visualize different endpoint structures and do not form a cross-dataset performance ranking or an integrated calibration efficacy result.}]{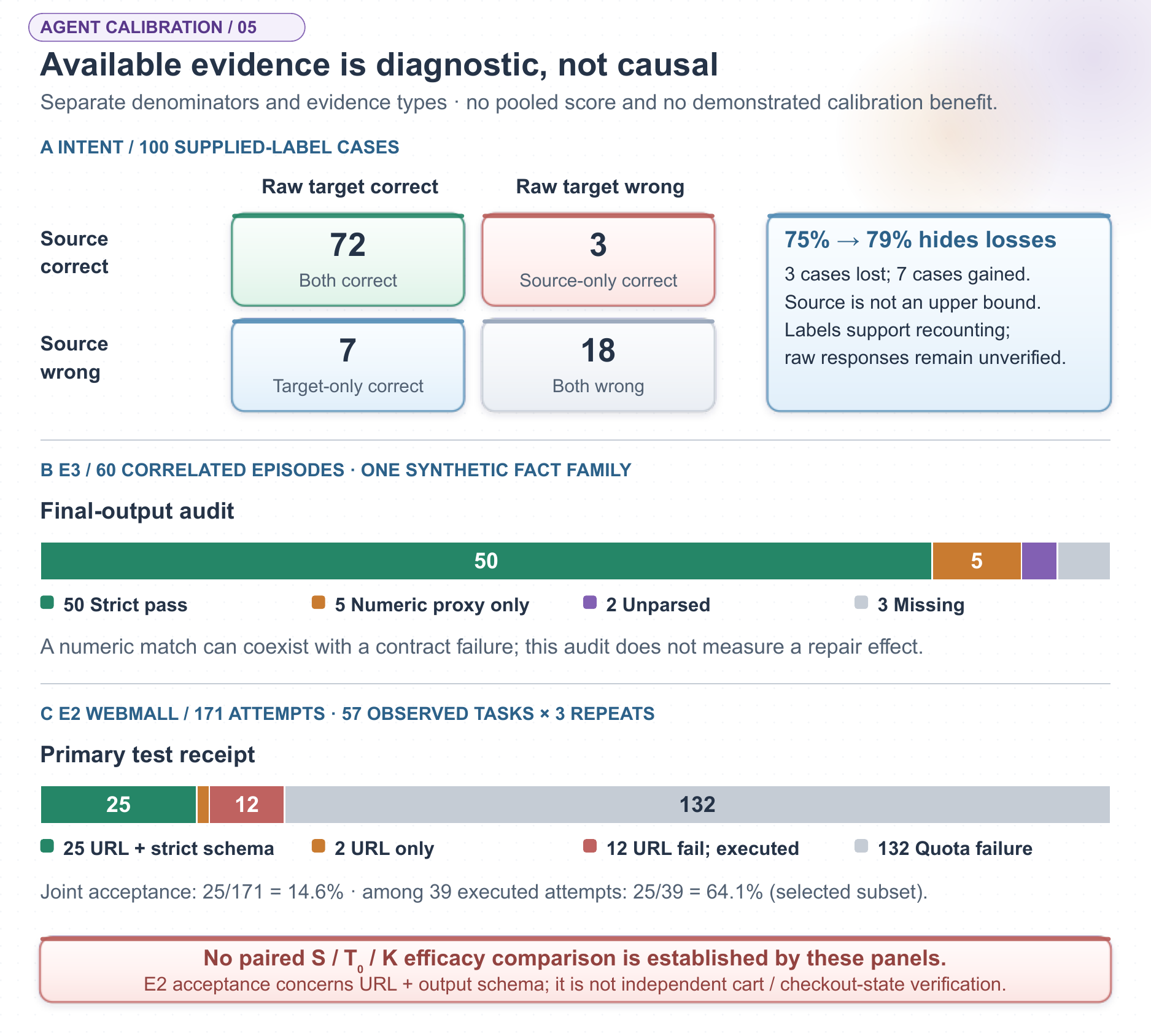}}
\caption{Diagnostic observations with separate denominators. Intent
counts are supplied-label outcomes for 100 items; E3 counts are 60
records from one synthetic family; WebMall counts are 171 target-only
attempts with 132 quota interruptions. Panels visualize different
endpoint structures and do not form a cross-dataset performance ranking
or an integrated calibration efficacy result.}
\end{figure}

\subsection{6.6. Evidence conclusion}\label{evidence-conclusion}

The current evidence supports four distinctions: replacement can produce
bidirectional differences; task proxies can disagree with artifact
contracts; availability changes the apparent success rate; and scoring
or provenance defects can masquerade as agent effects. It does not
establish integrated calibration efficacy, controller generalization,
user preference alignment, or lifecycle savings. An offline 27/27
pipeline check with zero foundation-model calls and a synthetic CPU
training prototype demonstrates limited pipeline feasibility, not
production RL. The next experiments must test the proposed mechanisms
rather than treating the existing diagnostics as their validation.

\FloatBarrier
\section{7. Prospective evaluation: how the hypotheses will be
tested}\label{prospective-evaluation-how-the-hypotheses-will-be-tested}

\subsection{7.1. Common protocol and
outcomes}\label{common-protocol-and-outcomes}

Freeze the contract, independent checker, splits, permitted edits, and
budget policy before optimization. Qualify real endpoints and meter
failures as well as successful runs. Partition by underlying question,
report family, template lineage, or environment instance; repeated
generations are nested observations. Separate calibration, development,
and previously unexposed test contexts. Use multiple model pairs and
both directions where resources permit, with recorded executor
identities. A no-op candidate remains eligible when \(T_0\) already
qualifies.

Primary outcomes are mandatory-suite qualification and joint destination
acceptance under a fixed contract. Secondary outcomes decompose basic,
technical, and user criteria; paired gains and regressions; induced
failures; valid abstentions and incomplete deliveries; and resources per
accepted delivery. Test-set independence and checker validity are
prerequisites, not optional reporting fields. Auditor assessment should
include known good and bad artifacts, injected violations, ambiguous
cases, and blinded expert adjudication where deterministic truth is
unavailable.

Uncertainty is computed at the independent sampling unit with repeats
nested within it. Select sample size using pilot variance and a
predeclared practically meaningful effect, not a universal three-repeat
rule or a noise floor borrowed from a different benchmark. Report
intervals and inconclusive outcomes. A source-retention result and a
destination-contract result can both be useful, but a changed contract
must not retroactively turn an old failure into a success.

For soft-quality preservation, predeclare a noninferiority margin; a
nonsignificant decline is not evidence of preservation. Following the
measurement concerns in {[}40{]}, qualify regression gates using user-
or report-family-clustered A/A splits and independently adjudicated
seeded unit, cross-section, and permission defects. Report false-alarm
rate and defect-detection power separately, including whether the
evidence covers the whole artifact.

Separate stopping the edit search from qualifying a release. Freeze one
candidate, the loss, the tolerance, and the sampling unit before opening
a fresh qualification set. If several candidates can win on that set,
prespecify the family and use valid risk tests with familywise error
control, following Learn then Test {[}73{]}. Repeated runs of one report
are dependent observations. A finite checklist can establish which
obligations passed on the observed cases. Extending that result to a
target population requires a defensible sampling model. A redesign
informed by qualification outcomes needs fresh qualification data.

A prerequisite checker study separates evaluation error from agent
error. Construct an expert-adjudicated artifact set containing valid
deliveries, factually correct but contract-invalid outputs, unsupported
confident answers, and permitted abstentions, together with answerable
controls. ImpossibleRubrics provides a precedent for pairing
evidence-insufficient cases with answerable controls {[}25{]}; its
supplied certificates and oracle audits still require validation for
this deployment. Freeze the independent evidence packet before exposing
cases to rubric generation or repair. Compare an outcome-only scorer, a
generic rubric scorer, and a contract-linked checker on the same
artifacts. Report false acceptance among adjudicated invalid artifacts,
false rejection among adjudicated valid artifacts, unresolved verdicts,
and unnecessary abstention on answerable tasks, each with its own
denominator. Then measure policy acceptance on separately held-out
tasks. Checker discrimination alone does not establish policy
improvement.

For the learning stage, vary the training verifier while holding the
qualification procedure fixed. Test both matched rollout budgets and
fully metered cost comparisons; cheaper grading may enable more rollouts
but does not guarantee lower total cost. Include rubric-keyword
repetition, polished unsupported claims, cross-language paraphrases, and
altered role permissions in a reserved stress set. Re-audit on newly
generated policy outputs to detect exploitation that a static artifact
set misses. A different judge family is a robustness check, not
sufficient independence: executable checks, separate reference evidence,
blinded adjudication, and exposure controls remain necessary. Report
reward gains jointly with qualification gains; disagreement is a result
to investigate, not a reason to replace the endpoint.

Under identical atomic criteria, compare flat compensatory, hard-gated,
and soft dependency-aware reward aggregation {[}49{]}. Scoring fixed
artifacts tests measurement behavior; to test calibration, optimize
separate candidates under each reward and evaluate them with the same
independent qualification procedure. Separately compare static and
adaptive development rubrics, logging additions, conflicts, coverage
loss, and review cost. Test semantically equivalent
qualification-checker variants against independent reference cases
{[}45{]}. For user context, compare correct, absent, stale, and swapped
preference histories while fixing authorized roles and permissions
{[}46{]}; then test authorized role changes separately. Hold out users
and time periods rather than only paraphrased requests.

\subsection{7.2. Mechanism comparisons}\label{mechanism-comparisons}

Include already-qualified cases and harmless rule-like patterns as
no-change controls. Report unsupported repair proposals per proposal,
unnecessary accepted edits per eligible no-change case, and regressions
induced by those edits. A higher reward with no improvement in the
applicable acceptance predicate is not sufficient evidence of a useful
repair.

Experiment A tests whether adaptation improves destination acceptance
beyond \(T_0\) and a strong target-native optimizer. Give methods the
same references, supervision access, editable components, and cost
limits. Report paired gains and losses, not only average scores.
Controlled professional changes include units, terminology, period
semantics, report-component dependencies, and authorized user roles. Raw
data versus approved estimated inputs must be distinct conditions.

The baseline set should include regularized harness optimization
{[}43{]} and atomic criteria with versioned aggregation, evaluator
audit, and component regression {[}40{]}. Then ablate explicit
Preserve/Transform/Add/Retire mapping while holding references,
criterion coverage, and optimization access constant. Label adapted
implementations accurately rather than claiming an exact reproduction.
This tests added value over established integration, not merely over an
unchanged prompt.

Experiment B tests the contribution of structured trajectories and
dependency-aware repair using the prospective arms in Table 6. These
names do not relabel historical T0 or C-lite. First hold the optimizer
and allowed action space constant. Give H0 and Htrace the same expert
facts, examples, and method content, provided as flat instructions in H0
and compiled into explicit checkpoints in Htrace; otherwise, added
information confounds the effect of structure. Meter annotation and
compilation labor separately. Then run a fixed-total-budget comparison,
where saved validation cost may fund additional search. The two designs
estimate different effects. Include a conservative full-revalidation
control and deliberately incomplete dependency graphs to expose unsafe
savings.

\begin{longtable}[]{@{}
  >{\raggedright\arraybackslash}p{(\linewidth - 4\tabcolsep) * \real{0.1593}}
  >{\raggedright\arraybackslash}p{(\linewidth - 4\tabcolsep) * \real{0.4613}}
  >{\raggedright\arraybackslash}p{(\linewidth - 4\tabcolsep) * \real{0.3794}}@{}}
\caption{Proposed, unrun intervention arms. Matched optimizer/access and
fixed-total-budget studies must be reported separately; behavior cloning
and fixed check/retry policies are required controller
controls.}\tabularnewline
\toprule\noalign{}
\begin{minipage}[b]{\linewidth}\raggedright
Prospective arm
\end{minipage} & \begin{minipage}[b]{\linewidth}\raggedright
Added mechanism
\end{minipage} & \begin{minipage}[b]{\linewidth}\raggedright
Main comparison
\end{minipage} \\
\midrule\noalign{}
\endfirsthead
\toprule\noalign{}
\begin{minipage}[b]{\linewidth}\raggedright
Prospective arm
\end{minipage} & \begin{minipage}[b]{\linewidth}\raggedright
Added mechanism
\end{minipage} & \begin{minipage}[b]{\linewidth}\raggedright
Main comparison
\end{minipage} \\
\midrule\noalign{}
\endhead
\bottomrule\noalign{}
\endlastfoot
H0 & Strong target-native harness with content-equivalent flat method
instructions & Baseline beyond historical raw T0 \\
Htrace & Same expert information compiled into method/checkpoint
structure & Increment from trajectory structure \\
Hdep & Htrace plus affected-dependency repair and revalidation &
Reliability and cost relative to full revalidation \\
Hctrl & Hdep plus a trained external decision policy & Increment beyond
strong untrained and supervised controls \\
\end{longtable}

Experiment C first crosses prompt-level configuration adaptation with
harness adaptation, yielding a strong target-native baseline B00 with
both configurations fixed, prompt only, harness only, and both. B00 is
distinct from the unchanged source-configured target \(T_0\) used for
migration diagnosis. For a common destination-acceptance endpoint \(q\),
estimate the interaction \(I_{12}=q_{11}-q_{10}-q_{01}+q_{00}\) with
paired, cluster-aware uncertainty. Positive, negligible, and negative
interactions are all possible. This estimates the effect of the declared
intervention packages, not universal independence between prompts and
tools. Hold access to references, shots, and memory constant, and report
both matched-action/access and equal-total-cost comparisons.

Then add the controller to each feasible base configuration, comparing
trained control with retrieval-only reuse, fixed routing/check/retry
policies, and behavior cloning. Freeze each cell's allowed action mask:
the controller cannot silently enable a prompt- or harness-level
capability disabled in that cell. If a controller necessarily changes
the action space, report a matched package increment rather than an
isolated factorial effect. Separate preference/RL improvement over
supervised training from the benefit of adding a controller at all. A
partial design identifies effects only for its tested bases; it cannot
claim all three-way interactions. Evaluate cleanly reloaded artifacts on
unseen task families, recipient templates, and user contexts. Offline
trace fit or unsupported off-policy estimates do not establish
performance under changed state distributions. For reusable skills,
qualify both the supervision and the produced API in the target
environment.

Keep an adaptation-time learned editor {[}41{]} separate from the
runtime-controller factor: edited harnesses can be deployed without
serving that editor on each request. Compare this editor as an
alternative harness-construction method before interpreting controller
gains. Process-credit methods adapted from {[}47, 48{]} require their
own ablation against outcome-only training of the same external policy
with the same frozen executor.

\textbf{Paired harness qualification.} Run the original and proposed
harness with both the source and replacement models on the same held-out
cases wherever all four combinations are executable. Record unavailable
combinations and their failed dependencies separately. Pin tool
responses for the diagnostic replay and repeat with live tools for
deployment qualification. Match the per-case budget and repetition count
across cells. For each model report paired counts of restored and newly
broken cases for each mandatory predicate, alongside the difference in
joint acceptance. Estimate uncertainty by resampling tasks with their
repeated runs kept together. The difference between the two within-model
repair effects measures how strongly the edit depends on the executor.
Charge edit search, failed probes and human review to the repair arm. If
acceptance is inconclusive, report the interval and cost per accepted
delivery; any claim of comparable acceptance requires a prespecified
tolerance. This comparison is prospective.

\textbf{Testing staged calibration.} Start with a report task in which a
known error in aggregation grain can be checked against input rows.
Include intact cases to measure needless edits. With the same frozen
executor, tools, edit proposals and total budget, vary one decision per
comparison: (1) phased priorities versus fixed equal weights, with a
Pareto candidate selector {[}5{]} as an additional comparator; (2) a
maintenance floor versus retaining the same criterion at low weight; (3)
requiring evidence that a diagnosed defect was repaired before tuning
versus permitting immediate tuning; and (4) expert-specified versus
model-proposed priority order, with the same information and review
allowance. All arms use identical final gates. Include unchanged and
generic-retry controls. Predeclare the floor, tolerance, repetition
count and budget before selecting candidates. On fresh paired tasks,
report joint acceptance, newly broken and restored cases,
protected-criterion regressions, unresolved runs, and cost per accepted
delivery including human review. Report same-case repair separately from
transfer to a new report. A lower search score or a costlier feasible
policy can still be the outcome.

To isolate the admission decision, replay each archived proposal against
its original incumbent. Give both rules the same visible runs. One rule
admits edits with sufficient net gain under a GUARD-inspired gain/loss
threshold {[}71{]}. The other also enforces the named contract gates and
floors. On independent cases, tabulate disagreements, acceptance with an
actual breach, rejection of a compliant improvement, and the cost of
retaining the incumbent. This comparison estimates a one-step decision
effect conditional on the proposal pool. Comparing complete search
policies requires a separate rollout because later candidates depend on
earlier admissions.

\subsection{7.3. Longitudinal value and stopping
decisions}\label{longitudinal-value-and-stopping-decisions}

Experiment D compares maintained harness and maintained harness plus
controller over a sequence of model, tool, template, user, and budget
changes. Separate changes to the input distribution from changes to the
acceptance contract. Both arms receive continued maintenance and the
same permissible information. Track initial build and training,
recurring annotation, engineering and review hours, GPU and API usage,
regressions, requalification, and accepted deliveries. Include cold
start and workload volumes insufficient to amortize training.

Meter rubric generation and verifier calls as well as agent execution.
AdaPRA's API-cost accounting illustrates that supervision can dominate
the bill {[}42{]}; RRSI's policy-token savings do not measure expert
labor {[}43{]}, and ORSD's matched environment-interaction budget does
not equal matched compute {[}48{]}. Report these resource categories
separately before computing total cost and accepted value.

\begin{figure}[!htbp]
\centering
\pandocbounded{\includegraphics[keepaspectratio,alt={Proposed lifecycle comparison. Experiment C crosses prompt and harness adaptation, then measures the controller increment on matched feasible bases. Both economic routes remain maintained and independently qualified over repeated deployment changes. The timeline corresponds to Experiment D and includes annotation, training, recurring labor, serving cost, and accepted delivery value. All quantities require measurement; the diagram asserts neither a break-even point nor superiority of training.}]{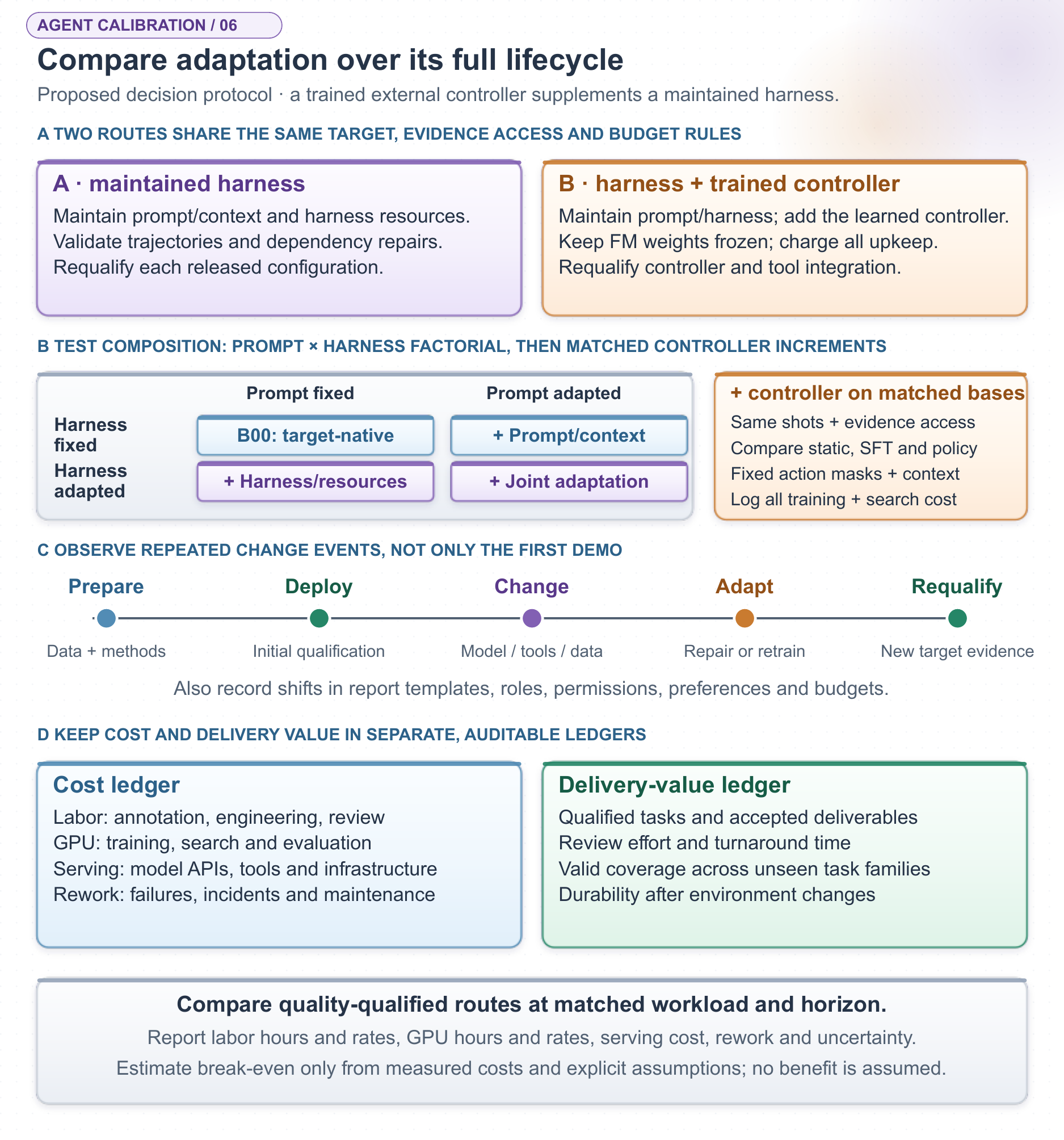}}
\caption{Proposed lifecycle comparison. Experiment C crosses prompt and
harness adaptation, then measures the controller increment on matched
feasible bases. Both economic routes remain maintained and independently
qualified over repeated deployment changes. The timeline corresponds to
Experiment D and includes annotation, training, recurring labor, serving
cost, and accepted delivery value. All quantities require measurement;
the diagram asserts neither a break-even point nor superiority of
training.}
\end{figure}

The decision is whether reusable policy learning improves accepted
output and net value under the same contract, not whether it produces a
higher training reward. Record reasons to stop: insufficient qualified
trajectories, unstable requirements, weak transfer, annotation costs
exceeding savings, or simpler deterministic fixes. Greater-than-80\%
zero-shot acceptance is a possible future deployment target, not a
current finding. Any such metric needs a defined population, frozen
policy, unseen-context protocol, and interval.

\subsection{7.4. Evidence ladder and staged
execution}\label{evidence-ladder-and-staged-execution}

The next study should first qualify the checker and provenance, then
test paired destination acceptance, then isolate trajectory and
dependency mechanisms, and only then evaluate learned transfer and
lifecycle value. A negative or inconclusive result at one stage is
retained rather than replaced by a different endpoint. Existing audits
occupy the diagnostic stage: they motivate these tests but cannot
substitute for any intervention comparison. The immediate priority is a
reproducible, modest-sized paired study with complete artifacts, not a
larger aggregate score built from mixed snapshots.

\subsection{7.5. Interactive elicitation and reusable
transfer}\label{interactive-elicitation-and-reusable-transfer}

Experiment E tests Section 4.5 prospectively. Compare manual prompt
editing, repeated sampling from a fixed prompt, an APPO-inspired
preference-guided prompt optimizer {[}55{]}, and configuration search
that additionally varies permitted evidence, plans, and tools. Add a
qualified-template reuse arm and a learned-controller arm only after
their training records exist. Hold task information, references,
candidate budgets, and allowed capabilities constant where isolating a
mechanism; report a separate equal-total-cost comparison. Record user
expertise and counterbalance condition order. A creative storyboard task
can test preference elicitation, while a professional report task with
checkable constraints tests whether preference gains coexist with
technical acceptance.

Add a fixed-configuration dueling-bandit baseline inspired by {[}56{]}
to distinguish choosing among known methods from generating new
configurations. Include an explicitly supplied-preference condition to
diagnose preference-inference failures, and replay verified tool outputs
in a diagnostic condition to isolate downstream execution failures
{[}58{]}; these privileged conditions are diagnostic ceilings, not
deployable competitors. Compare direct-choice-only labels against
inferred feedback labels, and compare outcome-only learning against
verified checkpoint supervision. Report preference identification,
execution qualification, and autonomous reuse separately. Split
retrospective training examples by underlying task or event before
constructing critiques, preventing variants of the same outcome from
entering both training and test sets.

Measure accepted deliveries, hard-rule violations, independent technical
correctness, user acceptance, rejection-of-all frequency, rounds, human
minutes, and generation/verification/training resources. Include
repeated samples, participant/task uncertainty intervals, and held-out
tasks, templates, users, and replacement models as distinct transfer
conditions. Freeze the reusable artifact before transfer testing. Test
whether gains survive without further user selection; do not combine
assisted session success with autonomous transfer. Audit position and
verbosity bias, preference drift, and unsupported checkpoint credit.
Better generalization and lower lifecycle cost are hypotheses, not
consequences of displaying more choices. None of the historical
diagnostics in Section 6 evaluates this mechanism.

Cross stable context with a declared role, audience, environment, or
budget shift. Compare context-conditioned memory against no memory and
context-blind pooled memory; report same-user repeated judgments,
disagreement, and preference reversals. Separate explicit authorized
changes from inferred preferences. Real recipient judgments support
user-acceptance claims; synthetic personas and simulated choices test
pipeline behavior only. Report context-specific results as well as
pooled averages so that improvement for one recipient cannot hide
regressions for another. Do not reuse the same person's repeated choices
as independent participants.

Add separate ablations for weight-conditioned candidate search,
evidence-linked reflection, and counter-probes. Compare mixed soft
objectives with fixed-weight and single-stream search under identical
hard gates and metered candidate budgets. Compare receipt-linked
reflection with an equally budgeted unstructured critique, measuring
unsupported check claims, unnecessary edits, independent acceptance, and
labor. Compare no probe, a budget-matched repeat or random admissible
contrast, and a hypothesis-directed counter-probe; an
uncertainty-directed clarification is a further active-elicitation
control. Use held-out judgments to test preference prediction and final
acceptance, rather than treating agreement with an inferred preference
as ground truth. Report probe burden and preference reversals, including
no response. An unseen-user claim requires disjoint user groups;
familiar users on new tasks form a separate protocol. Freeze the action
masks and reward/checker versions before each comparison.

For the profile update in Section 4.6, first replay a common,
independently assessed candidate set to compare fixed expert weights,
globally learned weights, and user/context-conditioned weights under
identical criteria, evidence and feedback budgets. This isolates
preference estimation from candidate quality. Then compare the resulting
generation policies on new tasks under the same allowed controls and
independent acceptance procedure. Report held-out preference prediction,
accepted deliveries and total cost separately. Treat changing rubric
content as a separate factor from fitting weights; test cohort priors on
disjoint users and distinguish cold-start from later personalized
sessions. Reward fit, predictive accuracy, and deployment improvement
need not move together.

These minimal controls do not replace comparisons with the closest
methods. Add a PARL-inspired personalized-rubric condition {[}66{]} in a
separate criterion-generation experiment, rather than attribute its
changed criteria to weight learning. Compare the proposed update gate
with a RubricWorld-inspired paired-edit acceptance rule {[}68{]}, using
the same edit proposals and certification budget. Report adaptations as
such, not as reproductions. A statistical acceptance claim requires
explicit sampling and label assumptions, sufficient certification data,
and an error budget across repeated edits; otherwise report empirical
acceptance only.

For each comparison, the evaluation record should identify dataset
provenance, split rules, trace fields, reviewer time, and GPU usage
(Appendix C). Proposed pilot sizes are feasibility choices, not
completed experiments or demonstrated statistical power.

\FloatBarrier
\section{8. Limitations and
implications}\label{limitations-and-implications}

This paper is a methodological proposal grounded in diagnostic audits.
Its integrated mechanism, automation, and economic benefits have not
been confirmed. The supplied studies have small or dependent samples,
incomplete provenance, mixed endpoints, and inconsistent controls.
Professional and user-context standards themselves require validation;
an explicit contract can still encode an incorrect rule, omit a
stakeholder, or contain conflicting requirements.

Hard gates can become brittle, and dependency models can omit
consequential links. Conservative revalidation reduces that risk at
additional cost. Expert trajectories can encode outdated practices or
organizational preferences that do not transfer. Temporal applicability,
reviewer disagreement, and permission boundaries therefore belong in the
supervision record. A compact policy can also overfit its annotations or
exploit reward defects. Frozen foundation weights do not make an agent's
behavior fixed or safe.

The practical implication is to organize adaptation around accepted use
rather than source imitation or average benchmark gains. Specify what
must remain valid and what must change; diagnose the gap with
independent evidence; choose a bounded harness or learning intervention;
and requalify the resulting deployment. The empirical question is
whether this discipline yields more accepted deliveries at an acceptable
lifecycle cost. The experiments above are designed to answer that
question, including the possibility that a simpler maintained harness is
sufficient.

\FloatBarrier
\section{References}\label{references}

\begin{enumerate}
\def\labelenumi{\arabic{enumi}.}
\item
  Yaxuan Wang, Quan Liu, Zhenting Wang, Zichao Li, Wei Wei, Yang Liu,
  and Yujia Bao. \textbf{PromptBridge: Cross-Model Prompt Transfer for
  Large Language Models.} 2025.
  \href{https://arxiv.org/abs/2512.01420}{arXiv:2512.01420}.
\item
  Yuhao Wu et al.~\textbf{HarnessDev: Can LLMs Create and Evolve Their
  Own Agent Harness?} 2026.
  \href{https://arxiv.org/abs/2609.01437}{arXiv:2609.01437}.
\item
  Varun Ursekar, Apaar Shanker, Yash Maurya, Shehab Yasser, Vijay S.
  Kalmath, Veronica Chatrath, and Yuan Xue. \textbf{HarnessOpt-Bench:
  Evaluating LLMs at Harness Optimization.} 2026.
  \href{https://arxiv.org/abs/2608.06301}{arXiv:2608.06301}.
\item
  Cen Mia Zhao, Haibo Ruan, Wenjie Chen, Pei-fen Tu, Usman Abbasi, and
  Joel Hesch. \textbf{Beyond Prompts: Measuring and Optimizing LLM
  Tool-Agent Harnesses.} 2026.
  \href{https://arxiv.org/abs/2609.05736}{arXiv:2609.05736}, v2,
  September 9 (first posted September 4).
\item
  Lakshya A. Agrawal et al.~\textbf{GEPA: Reflective Prompt Evolution
  Can Outperform Reinforcement Learning.} 2025; revised 2026.
  \href{https://arxiv.org/abs/2507.19457}{arXiv:2507.19457}.
\item
  Zibin Lin, Shengli Zhang, Taotao Wang, Yihan Xia, Deen Ma, and Guofu
  Liao. \textbf{Workflow-Localized Mechanism Learning:
  Attribution-Guided Repair and Knowledge Reuse for Structured Agent
  Skills.} 2026.
  \href{https://arxiv.org/abs/2607.20999v1}{arXiv:2607.20999v1}, July
  23.
\item
  Tejas Singh Anand et al.~\textbf{AEVAL: From Anecdotal to
  Deterministic Testing for Agentic Skill Workflows.} 2026.
  \href{https://arxiv.org/abs/2607.16345v2}{arXiv:2607.16345v2}, July 21
  (first posted July 16). Reviewed version: v2.
\item
  Yinghao Tang, Tan Zhenwei, Yiyao Wang, Wanli Gu, Xiaolu Zhang, Jun
  Zhou, and Wei Chen. \textbf{FinReportBench: Measuring and Improving
  Institution-Grade Financial Report Generation.} 2026.
  \href{https://arxiv.org/abs/2608.04374v1}{arXiv:2608.04374v1}, August
  5.
\item
  Haoyu Dong, Pengkun Zhang, Yan Gao, Xuanyu Dong, Yilin Cheng, Mingzhe
  Lu, Zikun Zhu, Adina Yakefu, and Shuxin Zheng. \textbf{Finch:
  Benchmarking Finance \& Accounting across Spreadsheet-Centric
  Enterprise Workflows.} 2026.
  \href{https://arxiv.org/abs/2512.13168v5}{arXiv:2512.13168v5}, April
  15; first posted December 15, 2025.
\item
  Wenjie Fu, Xiaoting Qin, Jue Zhang, Qingwei Lin, Lukas Wutschitz,
  Robert Sim, Saravan Rajmohan, and Dongmei Zhang. \textbf{CI-Work:
  Benchmarking Contextual Integrity in Enterprise LLM Agents.} ACL 2026
  Industry Track, pp.~1483--1508.
  \href{https://aclanthology.org/2026.acl-industry.103/}{ACL Anthology}.
  Conference July 2--7, 2026.
\item
  Kaiqu Liang et al.~\textbf{Learning Personalized Agents from Human
  Feedback.} 2026.
  \href{https://arxiv.org/abs/2602.16173}{arXiv:2602.16173}, v1,
  February 18.
\item
  Thi-Nhung Nguyen, Linhao Luo, Rollin Omari, Junae Kim, Thuy-Trang Vu,
  and Dinh Phung. \textbf{TriAlign: Towards Universal Truth Consistency
  in Personalized LLM Alignment.} 2026.
  \href{https://arxiv.org/abs/2606.01755v1}{arXiv:2606.01755v1}, June 1.
\item
  Haoyu Wang, Christopher M. Poskitt, and Jun Sun. \textbf{AgentSpec:
  Customizable Runtime Enforcement for Safe and Reliable LLM Agents.}
  ICSE 2026. \href{https://arxiv.org/abs/2503.18666}{arXiv:2503.18666},
  first posted March 24, 2025;
  \href{https://cposkitt.github.io/files/publications/agentspec_llm_enforcement_icse26.pdf}{author-hosted
  conference paper}.
\item
  Haiwen Yi and Xinyuan Song. \textbf{Learning to Control LLM Agent
  Harnesses with Offline Reinforcement Learning.} 2026.
  \href{https://arxiv.org/abs/2607.05458}{arXiv:2607.05458}, v1, July 5.
\item
  Mingkai Deng, Jianyu Wang, Cheng-Ping Hsieh, Yihan Wang, Han Guo,
  Tianmin Shu, Meng Song, Eric P. Xing, and Zhiting Hu.
  \textbf{RLPrompt: Optimizing Discrete Text Prompts with Reinforcement
  Learning.} EMNLP 2022, pp.~3369--3391.
  \href{https://aclanthology.org/2022.emnlp-main.222/}{ACL Anthology}.
\item
  Xufang Luo, Yuge Zhang, Zhiyuan He, Zilong Wang, Siyun Zhao, Dongsheng
  Li, Luna K. Qiu, and Yuqing Yang. \textbf{Agent Lightning: Train ANY
  AI Agents with Reinforcement Learning.} 2025.
  \href{https://arxiv.org/abs/2508.03680}{arXiv:2508.03680}.
\item
  Minki Kang, Jongwon Jeong, Seanie Lee, Jaewoong Cho, and Sung Ju
  Hwang. \textbf{Distilling LLM Agent into Small Models with Retrieval
  and Code Tools.} 2025.
  \href{https://arxiv.org/abs/2505.17612}{arXiv:2505.17612}.
\item
  Xue Bin Peng, Aviral Kumar, Grace Zhang, and Sergey Levine.
  \textbf{Advantage-Weighted Regression: Simple and Scalable Off-Policy
  Reinforcement Learning.} 2019.
  \href{https://arxiv.org/abs/1910.00177}{arXiv:1910.00177}, first
  posted October 1.
\item
  Rafael Rafailov, Archit Sharma, Eric Mitchell, Stefano Ermon,
  Christopher D. Manning, and Chelsea Finn. \textbf{Direct Preference
  Optimization: Your Language Model is Secretly a Reward Model.} 2023.
  \href{https://arxiv.org/abs/2305.18290}{arXiv:2305.18290}.
\item
  Zhihong Shao, Peiyi Wang, Qihao Zhu, Runxin Xu, Junxiao Song,
  Mingchuan Zhang, Y. K. Li, Y. Wu, and Daya Guo. \textbf{DeepSeekMath:
  Pushing the Limits of Mathematical Reasoning in Open Language Models.}
  2024. \href{https://arxiv.org/abs/2402.03300}{arXiv:2402.03300}.
\item
  Hunter Lightman et al.~\textbf{Let's Verify Step by Step.} 2023.
  \href{https://arxiv.org/abs/2305.20050}{arXiv:2305.20050}, May 31.
\item
  Anisha Gunjal, Anthony Wang, Elaine Lau, Vaskar Nath, Yunzhong He,
  Bing Liu, and Sean Hendryx. \textbf{Rubrics as Rewards: Reinforcement
  Learning Beyond Verifiable Domains.} 2025.
  \href{https://arxiv.org/abs/2507.17746}{arXiv:2507.17746}, v2, October
  3 (first posted July 23).
\item
  Weichu Xie et al.~\textbf{Step-wise Rubric Rewards for LLM Reasoning.}
  2026. \href{https://arxiv.org/abs/2605.17291}{arXiv:2605.17291}, v1,
  May 17.
\item
  Xuekang Wang, Zhuoyuan Hao, Shuo Hou, Hao Peng, Juanzi Li, and Xiaozhi
  Wang. \textbf{Reproducing, Analyzing, and Detecting Reward Hacking in
  Rubric-Based Reinforcement Learning.} 2026.
  \href{https://arxiv.org/abs/2606.04923}{arXiv:2606.04923}, v3,
  September 29 (first posted June 3).
\item
  Bowen Qin et al.~\textbf{ImpossibleRubrics: Stress-Testing Generated
  Rubrics as Reward Signals.} 2026.
  \href{https://arxiv.org/abs/2609.16816v1}{arXiv:2609.16816v1},
  September 15. Reviewed version: v1.
\item
  Sida Li, Yuanxin Liu, Shicheng Li, Linli Yao, Xu Sun, Haoli Bai, and
  Xiaohui Li. \textbf{Quality, Not Just Outcome: A Structural Scoring
  Framework for Selecting Coding-Agent Trajectories for Supervised
  Fine-Tuning.} 2026, October 9.
  \href{https://arxiv.org/abs/2610.13152}{arXiv:2610.13152}.
\item
  Kranti Chalamalasetti and Sowmya Vajjala. \textbf{LLM Judges Can Be
  Too Generous When There Is No Reference Answer.} 2026.
  \href{https://arxiv.org/abs/2607.12885}{arXiv:2607.12885}.
\item
  Umesh Bodhwani, Thanh Tran, and Kai Wei. \textbf{GAUGE: When Not to
  Trust LLM-as-a-Judge in User-Simulated Evaluation of Task-Oriented
  Agents.} 2026.
  \href{https://arxiv.org/abs/2609.12191}{arXiv:2609.12191}, v1,
  September 10.
\item
  Wenxiao Wang, Priyatham Kattakinda, and Soheil Feizi. \textbf{Do Agent
  Optimizers Compound? A Continual-Learning Evaluation on Terminal-Bench
  2.0.} 2026.
  \href{https://arxiv.org/abs/2607.14004v1}{arXiv:2607.14004v1}, July
  15. Reviewed version: v1.
\item
  Sushant Mehta et al.~\textbf{Cross-Benchmark Generalization in
  Long-Horizon Agents.} 2026.
  \href{https://arxiv.org/abs/2608.00181v1}{arXiv:2608.00181v1}, first
  posted July 31. Reviewed version: v1.
\item
  Eurostat / European Statistical System. \textbf{European Statistical
  System (ESS) guidelines on temporal disaggregation, benchmarking and
  reconciliation --- 2018 edition.} December 12, 2018. DOI:
  10.2785/743998.
  \href{https://ec.europa.eu/eurostat/web/products-manuals-and-guidelines/-/ks-06-18-355}{Official
  publication}.
\item
  Dan C. Hsu and Luke Lu. \textbf{Scoped Verification for Reliable
  Long-Horizon Agentic Context Evolution under Distribution Shift.}
  2026. \href{https://arxiv.org/abs/2607.09175v2}{arXiv:2607.09175v2},
  September 21; first posted July 10.
\item
  Su Wang et al.~\textbf{Phantom Guardrails: When Self-Improving Agent
  Harnesses Fix Failures That Never Happened.} 2026.
  \href{https://arxiv.org/abs/2607.13083v1}{arXiv:2607.13083v1}, July
  13.
\item
  Fei Wang and Zebai Tian. \textbf{Evidence-Bound Gateway-Path
  Provenance for Third-Party LLM Inference.} 2026.
  \href{https://arxiv.org/abs/2606.22560v2}{arXiv:2606.22560v2}, October
  16; first posted June 21.
\item
  Minye Shao et al.~\textbf{From Given to Gathered Evidence: Agentic
  Learning for Longitudinal Medical Reasoning.} 2026.
  \href{https://arxiv.org/abs/2609.39566v3}{arXiv:2609.39566v3}, October
  14; first posted September 30.
\item
  Zhou Yu, Bin Bi, Shiva Kumar Pentyala, Shubham Mehrotra, Sougata
  Chaudhuri, Shilpa Bhagavath, Zeyuan Chen, Ran Xu, Phil Mui, James Zhu,
  and Sitaram Asur. \textbf{Co-Evolving Harnesses and Models: On-Policy
  Correction Helps Weaker Models Catch Up Where Imitation Fails.} 2026.
  \href{https://arxiv.org/abs/2609.09134}{arXiv:2609.09134}, v1,
  September 8. Reviewed version: v1.
\item
  Andreas Plesner, Curtis Northcutt, Francisco Guzmán, and Anish
  Athalye. \textbf{A Cheap Verifier is Good Enough: LLM Post-training is
  Robust to Erroneous Rewards.} 2026.
  \href{https://arxiv.org/abs/2609.33467}{arXiv:2609.33467}, v1,
  September 27. Reviewed version: v1.
\item
  Hang He et al.~\textbf{WorkRover: Scaling Long-Horizon Cowork Agent
  Training Via Recursive Environment Evolution.} 2026.
  \href{https://arxiv.org/abs/2610.18067}{arXiv:2610.18067}, v1, October
  13. Reviewed version: v1.
\item
  Zifei Shan and Fangning Shao. \textbf{A Survey on Rubric-Guided
  Reinforcement Learning for Language Models.} 2026.
  \href{https://arxiv.org/abs/2608.27505v1}{arXiv:2608.27505v1}, August
  27. Reviewed version: v1.
\item
  Ruifeng Ren and Gun Ahn. \textbf{Closed-Loop Agent Evaluation: Layered
  Measurement and Self-Calibration.} 2026.
  \href{https://arxiv.org/abs/2611.07051v1}{arXiv:2611.07051v1},
  November 4.
\item
  Shuai Shao et al.~\textbf{Harness-R1: Learning to Edit Executable
  Runtime Harnesses from Agent Failure Trajectories.} 2026.
  \href{https://arxiv.org/abs/2608.02276v1}{arXiv:2608.02276v1}, August
  3.
\item
  Atharva Naik et al.~\textbf{AdaPRA: Process Reward Agents with
  Adaptive Rubrics for Data Science.} 2026.
  \href{https://arxiv.org/abs/2610.33245v1}{arXiv:2610.33245v1}, October
  23.
\item
  Peng Xia et al.~\textbf{RRSI: Regularized Recursive Self-Improvement
  of Agent Harnesses.} 2026.
  \href{https://arxiv.org/abs/2609.24972v3}{arXiv:2609.24972v3}, October
  4; first posted September 21.
\item
  Maoqi Liu et al.~\textbf{Scoring Higher, Answering Worse: Mitigating
  Reward Hacking in Rubric-Based RL via Protocol-Level Rubrics.} 2026.
  \href{https://arxiv.org/abs/2609.38847v1}{arXiv:2609.38847v1},
  September 30.
\item
  Wenbo Huang et al.~\textbf{Learning the Task or Learning the Verifier?
  Rethinking Criteria Exposure in Embodied Self-Improvement.} 2026.
  \href{https://arxiv.org/abs/2611.08318v1}{arXiv:2611.08318v1},
  November 5.
\item
  Garry Yang et al.~\textbf{APeB: Benchmarking Personalization Ability
  of Large Language Model Agents.} 2026.
  \href{https://arxiv.org/abs/2607.03162v3}{arXiv:2607.03162v3}, October
  16; first posted July 3.
\item
  Zihang Tian et al.~\textbf{ARCO: Adaptive Rubrics with Co-Evolution
  for Multi-Step LLM-Based Agents.} 2026.
  \href{https://arxiv.org/abs/2606.21262v2}{arXiv:2606.21262v2}, July
  23; first posted June 19.
\item
  Shumao Wu et al.~\textbf{Using Experience Twice: Dynamic Process
  Rubrics for Privileged Token-Level Credit Assignment.} 2026.
  \href{https://arxiv.org/abs/2611.06543v2}{arXiv:2611.06543v2},
  November 5; first posted November 4.
\item
  Can Lv, Mingju Chen, Heng Chang, and Shiji Zhou. \textbf{Mitigating
  False Credit Propagation: Probabilistic Graphical Reward Aggregation
  for Rubric-Based Reinforcement Learning.} 2026.
  \href{https://arxiv.org/abs/2606.03361v2}{arXiv:2606.03361v2}, August
  26; first posted June 2.
\item
  Son Nguyen, Xinyuan Liu, and Ransalu Senanayake. \textbf{CUPID in the
  Model Zoo: Online Matchmaking for Selecting Your Dream LLM.} 2026, May
  30. \href{https://arxiv.org/abs/2606.00846v1}{arXiv:2606.00846v1}.
\item
  Adam S. Jovine et al.~\textbf{LISTEN to Your Preferences: An LLM
  Framework for Multi-Objective Selection.} 2025; v3 May 30, 2026.
  \href{https://arxiv.org/abs/2510.25799v3}{arXiv:2510.25799v3}.
\item
  Mert Yuksekgonul et al.~\textbf{TextGrad: Automatic Differentiation
  via Text.} 2024, June 11.
  \href{https://arxiv.org/abs/2406.07496v1}{arXiv:2406.07496v1}.
\item
  Omar Khattab et al.~\textbf{DSPy: Compiling Declarative Language Model
  Calls into Self-Improving Pipelines.} 2023, October 5.
  \href{https://arxiv.org/abs/2310.03714v1}{arXiv:2310.03714v1}.
\item
  Zhijie Wang, Yuheng Huang, Da Song, Lei Ma, and Tianyi Zhang.
  \textbf{PromptCharm: Text-to-Image Generation through Multi-modal
  Prompting and Refinement.} CHI 2024; preprint March 6.
  \href{https://arxiv.org/abs/2403.04014v1}{arXiv:2403.04014v1}.
\item
  Zhipeng Li, Yi-Chi Liao, and Christian Holz. \textbf{Preference-Guided
  Prompt Optimization for Text-to-Image Generation.} CHI 2026; preprint
  February 13.
  \href{https://arxiv.org/abs/2602.13131v1}{arXiv:2602.13131v1}.
\item
  Romain Puech, Georgy Salakhutdinov, and Dimitris Bertsimas.
  \textbf{Online LLM Personalization from Pairwise Preferences Across
  Users.} 2026, October 12.
  \href{https://arxiv.org/abs/2610.17203v1}{arXiv:2610.17203v1}.
\item
  Jiamu Bai, Jiaming Hu, Yanhong Wu, and Zellux Wang.
  \textbf{Personalize at Test Time: Learning User Preferences for Image
  Generation.} 2026, October 6.
  \href{https://arxiv.org/abs/2610.09015v1}{arXiv:2610.09015v1}.
\item
  Dongchan Shin et al.~\textbf{AdaptArena: Evaluating Test-Time
  Personalization of Web Agents.} 2026, September 29.
  \href{https://arxiv.org/abs/2609.36488v1}{arXiv:2609.36488v1}.
\item
  Qiping Zhang, Kate Candon, Debasmita Ghose, and Marynel Vázquez.
  \textbf{Rethinking the Implications of Human Feedback for Preference
  Learning in Human-Robot Collaboration.} 2026, September 12.
  \href{https://arxiv.org/abs/2609.13982v1}{arXiv:2609.13982v1}.
\item
  Haoxiang Zhang et al.~\textbf{Self-Retrospection Distillation: Turning
  Post-hoc Experiences into Prior Foresight.} 2026; first posted October
  6, reviewed v2 October 8.
  \href{https://arxiv.org/abs/2610.08077v2}{arXiv:2610.08077v2}.
\item
  Hongliang Wei et al.~\textbf{HarnessBandit: Joint
  Learnability-Transferability Scheduling for Multi-Harness Agentic
  Reinforcement Learning.} 2026, September 12.
  \href{https://arxiv.org/abs/2609.13739v1}{arXiv:2609.13739v1}.
\item
  Constantinos Karouzos and Nikolaos Aletras. \textbf{Steerable
  Multi-Objective Group-Relative Preference Optimization with Jointly
  Trained Low-Rank Adapters.} 2026, November 6.
  \href{https://arxiv.org/abs/2611.09634v1}{arXiv:2611.09634v1}.
\item
  Duc Anh Vu, Thong Nguyen, Hoang Pham, Ponhvoan Srey, and Luu Anh Tuan.
  \textbf{Objective-Conditioned Active Preference Learning for
  Multi-Objective Alignment.} 2026, November 3.
  \href{https://arxiv.org/abs/2611.05506v1}{arXiv:2611.05506v1}.
\item
  Kandarp Joshi, Tushar Mittal, Chirag Soni, Lillian Zhou, and Petar
  Aleksic. \textbf{What's Worth Asking? Evaluating Preference
  Elicitation in Transactional Agents.} AACL-IJCNLP 2026, November
  6--10, pp.~3888--3914.
  \href{https://aclanthology.org/2026.aacl-long.174/}{ACL Anthology}.
\item
  Mengyuan Sun, Yu Li, Zhuohao Yu, Shikun Zhang, and Wei Ye.
  \textbf{Support Vector Rubrics: Closing the Gap Between Self-Generated
  and Human Rubrics.} EMNLP 2026, October 24--29, pp.~20226--20245.
  \href{https://aclanthology.org/2026.emnlp-main.875/}{ACL Anthology}.
\item
  Yilun Qiu, Xiaoyan Zhao, Yang Zhang, Yuxin Chen, Cilin Yan, Jiayin
  Cai, Xiaolong Jiang, Yao Hu, Yoko Yamakata, and Tat-Seng Chua.
  \textbf{Preference-Aware Rubric Learning for Personalized Evaluation.}
  2026, May 29.
  \href{https://arxiv.org/abs/2605.31545v1}{arXiv:2605.31545v1}.
\item
  Wei Chen, Yubing Wu, Shigui Li, Delu Zeng, Jingfeng Liu, Junmei Yang,
  Zhiheng Zhou, Bo Li, Min Chen, and Zhou Wang. \textbf{RubricOpt:
  Self-Revision of Semantic Rubrics for Preference Optimization.} 2026,
  October 21.
  \href{https://arxiv.org/abs/2610.29491v1}{arXiv:2610.29491v1}.
\item
  Hongji Pu, Hang Li, Yucheng Chu, and Jiliang Tang.
  \textbf{RubricWorld: World Model-Guided Rubric Optimization for LLM
  Graders.} 2026, November 1.
  \href{https://arxiv.org/abs/2611.01986v1}{arXiv:2611.01986v1}.
\item
  Ruiyao Xu et al.~\textbf{Harness\textsuperscript{2}: Recursive Agent Harnessing for an
  Open World.} 2026.
  \href{https://arxiv.org/abs/2610.14224v1}{arXiv:2610.14224v1}.
\item
  Yangze Liu and Zhongyi Han. \textbf{What Does a Harness Buy? Tokens,
  Mostly.} November 2026, v2.
  \href{https://arxiv.org/abs/2610.04433v2}{arXiv:2610.04433v2}.
\item
  Minghe Zhang, Xuetao Kang, Yifei Ming, Yanfei Chen, Han Yu, Yufan
  Zhuang, and Chen-Yu Lee. \textbf{GUARD: Stabilizing Recursive
  Self-Improvement with Regret Guidance.} November 2026.
  \href{https://arxiv.org/abs/2611.23476v1}{arXiv:2611.23476v1}.
\item
  Shaohuang Wang, Jinghao Wen, Xulei Zhang, and Xun Jiao.
  \textbf{Dependable Long-Horizon LLM Agents: A Trajectory-Centered
  Empirical Study of Where Errors Enter and Which Controls Reach Them.}
  November 2026.
  \href{https://arxiv.org/abs/2611.13254v1}{arXiv:2611.13254v1}.
\item
  Anastasios N. Angelopoulos, Stephen Bates, Emmanuel J. Candès, Michael
  I. Jordan, and Lihua Lei. \textbf{Learn then Test: Calibrating
  Predictive Algorithms to Achieve Risk Control.} The Annals of Applied
  Statistics 19(2), 1641--1662, 2025.
  \href{https://doi.org/10.1214/24-AOAS1998}{DOI}.
\end{enumerate}

\FloatBarrier
\section{Appendix A. Application profiles and observable
obligations}\label{appendix-a.-application-profiles-and-observable-obligations}

\textbf{QA: supported knowledge.} Build a reference inventory of
applicable claims, terminology, relations, provenance, and uncertainty.
Evaluate coverage, contradictions, unsupported additions, and recipient
comprehension. A semantic join must preserve entity identity, temporal
applicability, units, and source authority across inputs. Supported
target knowledge can improve the reference after authorized review; it
cannot silently redefine the test during optimization.

\textbf{Task: usable artifacts and actions.} Bind the task to recipient
templates, technical postconditions, and real endpoints. Changes in
search coverage or available tools require an approved evidence policy
and equivalent semantic capabilities, not necessarily identical APIs.
Sandboxes enforce access and resource boundaries; network or regional
partitions are implementation choices to verify, not evidence of
compliance by themselves.

\textbf{Job: reusable capabilities.} Qualify source and target traces
against the intended rewards and independent references before using
them as training data. Preserve useful complementary behavior without
treating either model as an oracle. Evaluate the produced controller or
skill on independent tasks, role boundaries, interface contracts, and
clean reload. Training convergence speed is empirical, and tests remain
necessary even if reward-driven learning is central.

\FloatBarrier
\section{Appendix B. Supplementary diagnostic
accounting}\label{appendix-b.-supplementary-diagnostic-accounting}

\subsection{B.1. FinQA endpoints and
resources}\label{b.1.-finqa-endpoints-and-resources}

\Needspace{12\baselineskip}
\begin{longtable}[]{@{}
  >{\raggedright\arraybackslash}p{(\linewidth - 10\tabcolsep) * \real{0.1632}}
  >{\raggedleft\arraybackslash}p{(\linewidth - 10\tabcolsep) * \real{0.1736}}
  >{\raggedleft\arraybackslash}p{(\linewidth - 10\tabcolsep) * \real{0.1632}}
  >{\raggedleft\arraybackslash}p{(\linewidth - 10\tabcolsep) * \real{0.1736}}
  >{\raggedleft\arraybackslash}p{(\linewidth - 10\tabcolsep) * \real{0.1632}}
  >{\raggedleft\arraybackslash}p{(\linewidth - 10\tabcolsep) * \real{0.1632}}@{}}
\caption{FinQA R1 supplied aggregates, 300 episodes per arm from 100
questions and three repeats. Execution and strict success are not
interchangeable.}\tabularnewline
\toprule\noalign{}
\begin{minipage}[b]{\linewidth}\raggedright
Arm
\end{minipage} & \begin{minipage}[b]{\linewidth}\raggedleft
Execution \%
\end{minipage} & \begin{minipage}[b]{\linewidth}\raggedleft
Answer \%
\end{minipage} & \begin{minipage}[b]{\linewidth}\raggedleft
Structure \%
\end{minipage} & \begin{minipage}[b]{\linewidth}\raggedleft
Strict \%
\end{minipage} & \begin{minipage}[b]{\linewidth}\raggedleft
Mean calls
\end{minipage} \\
\midrule\noalign{}
\endfirsthead
\toprule\noalign{}
\begin{minipage}[b]{\linewidth}\raggedright
Arm
\end{minipage} & \begin{minipage}[b]{\linewidth}\raggedleft
Execution \%
\end{minipage} & \begin{minipage}[b]{\linewidth}\raggedleft
Answer \%
\end{minipage} & \begin{minipage}[b]{\linewidth}\raggedleft
Structure \%
\end{minipage} & \begin{minipage}[b]{\linewidth}\raggedleft
Strict \%
\end{minipage} & \begin{minipage}[b]{\linewidth}\raggedleft
Mean calls
\end{minipage} \\
\midrule\noalign{}
\endhead
\bottomrule\noalign{}
\endlastfoot
S & 65.0 & 63.7 & 98.3 & 63.7 & 2.00 \\
T0 & 63.3 & 58.3 & 90.7 & 58.0 & 2.00 \\
T2 & 59.3 & 59.0 & 89.3 & 57.3 & 2.30 \\
C-lite & 62.7 & 59.0 & 94.7 & 59.0 & 2.17 \\
\end{longtable}

\begin{longtable}[]{@{}
  >{\raggedright\arraybackslash}p{(\linewidth - 10\tabcolsep) * \real{0.1632}}
  >{\raggedleft\arraybackslash}p{(\linewidth - 10\tabcolsep) * \real{0.1736}}
  >{\raggedleft\arraybackslash}p{(\linewidth - 10\tabcolsep) * \real{0.1632}}
  >{\raggedleft\arraybackslash}p{(\linewidth - 10\tabcolsep) * \real{0.1736}}
  >{\raggedleft\arraybackslash}p{(\linewidth - 10\tabcolsep) * \real{0.1632}}
  >{\raggedleft\arraybackslash}p{(\linewidth - 10\tabcolsep) * \real{0.1632}}@{}}
\caption{FinQA R2 supplied aggregates, 100 episodes per arm. Distinct
directions and repeat structures preclude pooling with R1 as independent
questions.}\tabularnewline
\toprule\noalign{}
\begin{minipage}[b]{\linewidth}\raggedright
Arm
\end{minipage} & \begin{minipage}[b]{\linewidth}\raggedleft
Execution \%
\end{minipage} & \begin{minipage}[b]{\linewidth}\raggedleft
Answer \%
\end{minipage} & \begin{minipage}[b]{\linewidth}\raggedleft
Structure \%
\end{minipage} & \begin{minipage}[b]{\linewidth}\raggedleft
Strict \%
\end{minipage} & \begin{minipage}[b]{\linewidth}\raggedleft
Mean calls
\end{minipage} \\
\midrule\noalign{}
\endfirsthead
\toprule\noalign{}
\begin{minipage}[b]{\linewidth}\raggedright
Arm
\end{minipage} & \begin{minipage}[b]{\linewidth}\raggedleft
Execution \%
\end{minipage} & \begin{minipage}[b]{\linewidth}\raggedleft
Answer \%
\end{minipage} & \begin{minipage}[b]{\linewidth}\raggedleft
Structure \%
\end{minipage} & \begin{minipage}[b]{\linewidth}\raggedleft
Strict \%
\end{minipage} & \begin{minipage}[b]{\linewidth}\raggedleft
Mean calls
\end{minipage} \\
\midrule\noalign{}
\endhead
\bottomrule\noalign{}
\endlastfoot
S & 64 & 63 & 98 & 63 & 2.00 \\
T0 & 60 & 58 & 91 & 58 & 2.00 \\
T2 & 58 & 58 & 96 & 58 & 2.30 \\
C-lite & 60 & 60 & 91 & 59 & 2.22 \\
\end{longtable}

R2 mean summed API time is reported as 19.5, 60.4, 199.8, and 411.2
seconds for S, T0, T2, and C-lite. This is not end-to-end latency, and
the C-lite maximum of 32,098 seconds remains unresolved. Known C-lite
tokens are 600,785 versus T0's 505,816, approximately 18.8\% more, with
25 versus ten missing usage records; T2 has 23 missing records. These
incomplete measurements cannot establish cost efficiency. One aggregate
ZIP contains five distinct members named summary.json, including
different calibration/DEV/Test variants; these require ordinal-qualified
provenance and must not be silently merged. A manifest is empty, model
names conflict across metadata and logs, and an 80-row review worksheet
covers only 20 IDs with blank reviewer identities. It is not an
independent user study.

\subsection{B.2. Other intent, financial, and environment
records}\label{b.2.-other-intent-financial-and-environment-records}

The separate Intent A-to-B summary reports accuracy S/T0/T1/T2/C-lite of
79/77/76/80/77 and structure rates 100/94/90/97/94. B-to-A source
structure is 93\%, distinct from its 75\% label accuracy. These
summaries lack the controls needed to infer a direction-specific model
advantage.

For E3, some identity checks pass vacuously on unparsed or empty
content; the 57 identity passes are not 57 validated artifacts. The
per-arm Test denominator is six repeated records within one synthetic
family. A later E3 summary describes additional planned and partial runs
but has unresolved paired-table and denominator conflicts; it is not
merged into the 60-record audit.

For WebMall, the calibration snapshot has 60 attempts: 43 joint URL/JSON
passes, four URL-only passes, and 13 executed nonpasses. DEV has 33
attempts: 21 joint passes, eight URL-only passes, and four executed
nonpasses. Neither contains quota interruptions. Reported 17/20 and
10/11 phase success values are best-of-three task rates, unlike these
attempt rates. Test's 25/2/12/132 categories are joint pass, URL-only
pass, executed nonpass, and quota interruption. Its 55.82-second
all-attempt mean and roughly 233-second executed-attempt mean use
different denominators.

A separate progress report describes T0 23/39 executed successes; S five
successes among nine executed or 36 total attempts, including 27
quota/timeouts; and one successful calibration smoke test. T1 was unrun.
Raw matched traces are absent, and an offline check still fails the
smoke output's K1 criterion. Re-scoring 19 old outputs from 10 to 12
passes is an evaluator correction, not agent improvement. Token coverage
is only 32/76 attempts. Repair rounds and reported tokens are not RL
training evidence.

Five catalog CSVs contain 110 import rows, not 110 independent products
or agent episodes. Two 26-row matched pairs change every SKU; image
fields change in eight and one rows respectively, while matched IDs,
names, prices, and descriptions remain stable. Chronology and causal
effects on failures are unverified. These assets motivate version
binding and environment reconstruction rather than a new efficacy claim.

\FloatBarrier
\section{Appendix C. Minimum reproducible
record}\label{appendix-c.-minimum-reproducible-record}

A deployable calibration package should record the contract and scope
fingerprint; source-to-target requirement mapping; reference provenance
and temporal applicability; method graph and observed trace IDs;
checkpoint verdicts with evidence and uncertainty; dependencies and
invalidation sets; absolute reward anchors and justified relative pairs;
candidate edits or training artifacts; split membership and exposure
history; checker identity and independent audit; all attempted runs,
including missing and interrupted ones; complete cost meters and
missingness; and a bounded release or escalation decision.

The legacy studies do not yet satisfy this proposed schema. Historical
manuscripts and audits remain archived; future controlled results must
add evidence without rewriting earlier observations.

\FloatBarrier
\section{Appendix D. Contract-to-feedback
interfaces}\label{appendix-d.-contract-to-feedback-interfaces}

\begin{longtable}[]{@{}
  >{\raggedright\arraybackslash}p{(\linewidth - 4\tabcolsep) * \real{0.1606}}
  >{\raggedright\arraybackslash}p{(\linewidth - 4\tabcolsep) * \real{0.4296}}
  >{\raggedright\arraybackslash}p{(\linewidth - 4\tabcolsep) * \real{0.4098}}@{}}
\caption{Proposed interfaces connecting expert standards to optimization
and qualification. These are design requirements, not a validated
implementation.}\tabularnewline
\toprule\noalign{}
\begin{minipage}[b]{\linewidth}\raggedright
Object
\end{minipage} & \begin{minipage}[b]{\linewidth}\raggedright
Minimum recorded fields
\end{minipage} & \begin{minipage}[b]{\linewidth}\raggedright
Boundary
\end{minipage} \\
\midrule\noalign{}
\endfirsthead
\toprule\noalign{}
\begin{minipage}[b]{\linewidth}\raggedright
Object
\end{minipage} & \begin{minipage}[b]{\linewidth}\raggedright
Minimum recorded fields
\end{minipage} & \begin{minipage}[b]{\linewidth}\raggedright
Boundary
\end{minipage} \\
\midrule\noalign{}
\endhead
\bottomrule\noalign{}
\endlastfoot
Acceptance contract & Requirement ID, owner, scope, effective version,
mandatory predicate, authorized exceptions & Only authorized requirement
changes alter acceptance \\
Task/checkpoint rubric & Linked requirement, applicability, expected
evidence, anchors, dependent artifacts & Criterion coverage is reviewed
before use; a heading is not semantic evidence \\
Verification record & Criterion ID, runtime receipt, before/after
artifact and evidence IDs, comparator and checker/version, verdict,
uncertainty & Missing execution and insufficient evidence remain
distinct; a rationale is not a receipt \\
Optimization feedback & Separate anchored/contextual scores, scale
anchors and weights, justified pair, provenance, cost meters & Preserve
raw observations; adjustable soft rewards cannot compensate for
mandatory violations \\
Preference profile & Rubric/scale version, user/context scope, eligible
feedback, weights with uncertainty, update rule and rollback & Profile
inference neither changes a hard predicate nor qualifies an executable
configuration \\
Qualification record & Frozen candidate and contract, independent
checker/reference, split exposure, scoped decision & A training score
cannot authorize release \\
\end{longtable}

The mapping is many-to-many: one checkpoint may assess technical and
user obligations, and a requirement may span several report components.
Checkpoint observations are proposed credit-assignment signals rather
than proven causal contributions to the final outcome. The registry
retains raw verdicts even when an optimizer requires a scalar. Shared
evidence across criteria must be recorded to avoid treating correlated
scores as independent support. A contract revision triggers a coverage
review of the rubric, regression cases for the verifier, and
requalification of affected deployment scopes.

\end{document}